\documentclass{article}

\usepackage{graphicx}
\usepackage{subcaption}
\usepackage{wrapfig}
\usepackage[preprint]{nips_2018}

\usepackage{amsthm}
\usepackage{algorithm}
\usepackage{algorithmic}
\usepackage[utf8]{inputenc} 
\usepackage[T1]{fontenc}    
\usepackage{hyperref}       
\usepackage{url}            
\usepackage{booktabs}       
\usepackage{amsfonts}       
\usepackage{nicefrac}       
\usepackage{microtype}      
\usepackage{amsmath}
\usepackage{amssymb}
\usepackage{multirow}
\usepackage{bm}
\newtheorem{theorem}{Theorem}[section]
\newtheorem{lemma}[theorem]{Lemma}
\newtheorem{proposition}[theorem]{Proposition}

\newtheorem{assumption}{Assumption}
\newtheorem{definition}{Definition}
\newtheorem{fact}[theorem]{Fact}
\newcommand{\A}{\mathbf{A}}

\newcommand{\D}{\mathbf{D}}

\newcommand{\e}{\mathbf{e}}

\newcommand{\I}{\mathbf{I}}

\renewcommand{\v}{\mathbf{v}}

\newcommand{\x}{\mathbf{x}}

\newcommand{\y}{\mathbf{y}}

\newcommand{\bmu}{\bm{\mu}}

\title{Adaptive Stochastic Gradient Langevin Dynamics: Taming Convergence and Saddle Point Escape Time}

\author{
  Hejian Sang
\\
  Department of Statistics\\
  Iowa State University\\
  Ames, IA, 50011 \\
  \texttt{hjsang@iastate.edu}
  \And
  Jia Liu\\
  Department of Computer Science\\
   Iowa State University\\
  Ames, IA, 50011 \\
  \texttt{jialiu@iastate.edu} \\
}

\begin{document}

\maketitle

\begin{abstract}
	In this paper, we propose a new adaptive stochastic gradient Langevin dynamics (ASGLD) algorithmic framework and its two specialized versions, namely adaptive stochastic gradient (ASG) and adaptive gradient Langevin dynamics(AGLD), for non-convex optimization problems.
	All proposed algorithms can escape from saddle points with at most $O(\log d)$ iterations, which is nearly dimension-free. 
	Further, we show that ASGLD and ASG converge to a local minimum with at most $O(\log d/\epsilon^4)$ iterations. 
	Also, ASGLD with full gradients or ASGLD with a slowly linearly increasing batch size converge to a local minimum with iterations bounded by $O(\log d/\epsilon^2)$, 
	which outperforms existing first-order methods.
\end{abstract}

\section{Introduction}\label{sec::intro}

{\bf Background and motivation:}
Since its inception as a discipline, machine learning has made an extensive use of optimization techniques and algorithms.
At the heart of many machine learning applications usually lies a minimization problem for some non-convex function $f(\x)$ in the form of: 
\begin{align} \label{eqn_orig_formulation}
\min_{\x \in \mathbb{R}^{d}} f(\x) \triangleq \min_{\x \in \mathbb{R}^{d}} \frac{1}{n} \sum_{i=1}^{n} F(\x,\xi_{i}),  
\end{align}
where $F(\x,\xi_{i})$ is a random smooth non-convex function of $\x \in \mathbb{R}^{d}$ that depends on random data samples $\xi_{i}$, $i=1,\ldots,n$.
The problem formulation in (\ref{eqn_orig_formulation}) frequently arises in many deep learning problems and empirical loss minimization in statistical applications (e.g., \cite{goodfellow2016deep}).
Exacerbating the problem is the fact that, in many learning applications, the problem dimension $d$ and the sample space size $n$ are usually large (see, e.g., \cite{natarajan1995sparse,candes2009exact,candes2015phase}).
The high dimensionality and large sample space imply that computing and storing a high-order oracle (e.g., Hessian matrix) is prohibitively expensive and it is only feasible to employ simple first-order stochastic gradient descent (SGD) algorithm or its variants (see, e.g., \cite{lin2007projected,chen2015fast,dempster1977maximum,mclachlan2007algorithm,zhang2017hitting,ge2015escaping,raginsky2017non}, to name but a few).

Simply speaking, the SGD algorithm takes the following iterative form: $\x_{k+1} = \x_{k} - \eta_{k} G(\x_{k})$, where $G(\x_{k})$ is a stochastic gradient at $\x_{k}$ and $\eta_{k}$ represents the step-size in the $k$-th iteration. 
Due to the NP-hard nature of non-convex problems \citep{blum1989training,meka2008rank}, finding a local minimum is usually the best hope and also a desired goal in non-convex learning\footnote{
	There exists a vast literature on probabilistically finding global minimum by identifying local minima, e.g., graduated optimization \citep{hazan2016graduated, gashler2008iterative} and random seeding \citep{zabinsky2009random}.
	Moreover, in some special-structured non-convex problems (e.g., \cite{Laurent17:LocalGlobal,Lu17:LocalGlobal,Kawaguchi16:LocalGlobal}, etc.), it can be shown that every local minimum is a global minimum.}.
Therefore, how to develop an SGD-type algorithm to efficiently converge to a local minimum is of central interest in non-convex learning. On the other hand, however, due to its first-order nature, an SGD-type algorithm could easily be trapped at an $\epsilon$-saddle point (i.e., a stationary point $\x$ with $\|f(\x)\| \leq \epsilon$). Therefore, one also wants to design an SGD-type algorithm that can quickly escape from a saddle point. However, these two important goals (i.e., converging to a local minimum efficiently and escaping from a saddle point quickly) are {\em conflicting} by nature, and how to reconcile them has become one of the major challenges in non-convex learning.


To facilitate saddle point escape, stochastic gradient Langevin dynamics (SGLD) has recently emerged as a promising solution \citep{welling2011bayesian,neelakantan2015adding,kaiser2015neural,teh2016consistency}. The basic idea of SGLD is to inject a {\em structured} random noise based on the so-called Langevin dynamics (LD) into the basic SGD scheme, which enables the escape from a saddle point. More precisely, the traditional SGLD algorithm can be written as:
\begin{align} \label{sgld}
\x_{k+1}=\x_{k}-\eta G(\x_{k})+\sqrt{2\eta/u}\e_k, \quad k=1,2,3,\ldots,
\end{align}
where $\e_{k}$ is a zero-mean Gaussian random vector independent of $\x_{k}$ and $u>0$ is a temperature parameter.
In \citep{zhang2017hitting}, the hitting time of SGLD is analyzed.
Moreover, \cite{teh2016consistency}  \cite{raginsky2017non}, and \cite{xu2017global} studied the convergence of SGLD (see more detailed discussions in Section \ref{sec::related_work}).  
However, the use of SGLD naturally raise the question on how to strike a good balance between the magnitudes of $G(\x_{k})$ and the LD-based random noise $\sqrt{2\eta/u} \e_{k}$.
If not treated appropriately, this imbalance could negatively affect the convergence rate and the escape capability of SGLD.  
Another major limitation of SGLD is that the escape time from saddle points grows polynomially in $d$ \citep{zhang2017hitting}.
Hence, identifying a saddle point escape direction with SGLD in a high-dimensional space is, in some sense, analogous to finding a needle in a haystack. 
Therefore, there is a compelling need for resolving this curse of dimensionality for SGLD.
Moreover, in SGLD, the step size and temperature parameter need to be chosen carefully to make the algorithm work efficiently. 
So far, however, there is no unified way to {\em systematically} tune these parameters.
All of these theoretical concerns motivate us to propose a new adaptive SGLD (ASGLD) algorithmic framework, which guarantees better convergence and escape performances with lower complexity in tuning the hyper-parameters.

{\bf Our approach and contributions:}
Motivated by the adaptive stochastic gradient method (ASG) in \citep{duchi2011adaptive}, our proposed ASGLD framework uses the historical information of stochastic gradients to adaptively guide and adjust the search direction and LD-based random noise. The rationale behind this adaptation is to i) simplify setting step size, ii) avoid explicitly specifying temperature parameter in SGLD, and most importantly, iii) incorporate the geometry of the objective function, which gives frequently occurring features low learning rates and infrequent features high learning rates.
Our proposed ASGLD algorithmic framework can be written as follows:
\begin{align} \label{asgld}
\x_{k+1}=\x_{k}-\eta \D_{k-1}^{-\alpha/2} G(\x_{k})+\sqrt{2\eta}\D_{k-1}^{-\beta/2}\e_{k},
\end{align}
where $\D_{k-1} \triangleq \sum_{i=1}^{k-1} \mathrm{Diag}\{ G(\x_{i}) \star G(\x_{i}) + \delta \I_{d\times d}\}$.
Here, ``$\star$'' represents the element wise product and $\delta>0$ is a constant. 
In ASGLD, the balance between stochastic gradient $G(\x_{k})$ and the LD-based noise is governed by the 2-tuple $(\alpha, \beta)$. 
A salient feature of ASGLD is that one has the flexibility to choose the most suitable $(\alpha,\beta)$ combination, depending on whether convergence or escape time is {\em more favored} in his/her performance needs.
In this paper, our goal is to understand and characterize the convergence and saddle point escape time of ASGLD with all possible combinations of $(\alpha, \beta)$.

Moreover, our ASGLD algorithm is a unifying framework that includes SGLD ($\alpha=0,\beta = 0$), ASG ($\alpha =1,\beta=\infty$), and SGD ($\alpha=0,\beta=\infty$) as special cases. 
However, compared to SGLD and ASG, the introduction of the diagonal matrix $\D_{k}$ and its powers of $\alpha$ and $\beta$ significantly complicate the convergence and saddle point escape time analysis of ASGLD and necessitate new proof techniques.
Our main contribution in this paper is that we develop a series of non-trivial techniques to handle the adaptive matrix term $\D_{k-1}$ in ASGLD and establish the superior convergence and escape time performance of ASGLD.
The main technical results in this paper are summerized as follows:


\begin{list}{\labelitemi}{\leftmargin=1em}

	\item We propose a unifying ASGLD algorithmic framework governed by two control degrees of freedom $(\alpha, \beta)$. 
	These two control parameters allow us to avoid directly specifying the inverse temperature parameter as in SGLD and simply use a constant step size is to establish its convergence.
	Further, ASGLD provides a unifying framework that covers all variants of LD-based SGD, including SGD, SGLD, and ASG, as well as its deterministic version AGLD.
	We first establish the convergence of the most general ASGLD framework to a first-order stationary point, and then specialize it to AGLD and ASG. 
	We show that ASGLD converges for $\alpha\in(0,2]$ and $\beta>\alpha/2$.
	This result subsequently implies that the convergence of AGLD is guaranteed for $\alpha\in[0,2]$ and $\beta>\alpha/2$ and that ASG converges for $\alpha\in(0,2]$.
	
	\item In addition to establishing convergence, we further characterize the convergence rates of ASGLD and its specialized versions.
	Our convergence rate analysis shows that ASGLD achieves the fastest convergence time when the magnitudes of stochastic gradient and LD-based injected noise are perfectly balanced (i.e., $\alpha=\beta=1$), where the best convergence time complexity is $\tilde O(1/\epsilon^4)$. 
	ASG also achieves the best convergence time $\tilde O(1/\epsilon^4)$ when $\alpha=1$. 
	If full gradients are available, the deterministic counterpart AGLD further improves the best convergence time to $\tilde O(1/\epsilon^2)$ (achieved by $\alpha=0$ and $\beta=1$). 
	Notably, using a slowly linearly increasing mini-batch with batch size growing at $O(k)$, ASGLD achieves the best convergence time $\tilde{O}(1/\epsilon^2)$.
	
	\item  Last by not least, we characterize the saddle point escape time for the general ASGLD and its two specialized versions AGLD and ASG. 
	We show that our proposed ASGLD framework enjoys an escape time that only grows at rate $O(\log d)$, which is  almost {\em free of dimension} and significantly sharper than the state-of-the-art $O(\log^4 d)$ \citep{jin2017escape}.
	Specifically, we show that ASGLD with constant batch size and ASG converge to local minimum with at most $\tilde O(\log d/\epsilon^4)$ iterations, while at most $\tilde O(\log d/\epsilon^2)$ iterations are needed to guarantee local minimum for AGLD and ASGLD with a slowing linearly increasing batch size. 
	To our knowledge, this is the fastest saddle point escape time in existing first-order methods and even competitive compared to some second-order methods (see Section~\ref{sec::related_work} for more detailed discussions).
\end{list}

The remainder of the paper is organized as follows:
Section~\ref{sec::related_work} puts ASGLD into comparative perspectives.
Section~\ref{sec::algorithm} presents the general ASGLD framework and its specialized versions.
Section~\ref{theory} summarizes all main theoretical results, and Section~\ref{sec::conclusion} concludes the paper.
Due to space limitation, we relegate most of the proofs and numerical results to supplementary material of this paper.


\section{Related work}\label{sec::related_work}

In this section, we provide an overview on saddle point escape for non-convex learning, which has attracted a lot of attention in the machine learning community recently.
Generally speaking, the most straightforward way of escaping from a saddle point is to examine the second-order Hessian matrix and identify an eigen-direction associated with a negative eigenvalue as an escape direction.
Not surprisingly, most early attempts in this area are based on second-order curvature information.
For example, \cite{nesterov2006cubic} used a cubic regularization with Hessian information to obtain dimension-free $O(1/\epsilon^{1.5})$ time complexity to escape saddle points. 
The same time complexity was obtained by \cite{curtis2017trust} by using a second-order trust region algorithm.  
Later, escape time complexity based on second-order approach was sharpened to $O(\log d/\epsilon^{1.25})$ by \cite{carmon2016accelerated}, where only Hessian-vector product information is needed (hence an $O(\log d)$ penalty is incurred).
\cite{agarwal2017finding} also proposed a cubic regularization method based on \cite{cartis2011adaptive} to achieve the same $O(\log d/\epsilon^{1.25})$ escape time.

As mentioned earlier, due to the high-dimensionality in many learning problems, acquiring Hessian information is too expensive and saddle point escape based only on first-order information is much more preferred.
To this end, \cite{ge2015escaping} proposed an SGD algorithm with injected sphere noise, where  the escape time complexity depends polynomially on $d$ (the polynomial order is at least four). 
\cite{lee2016gradient} demonstrated that gradient descent with random initialization almost surely converges to a local minimum. 
However, the escape time from saddle point is not studied.  
\cite{xu2017first} extracted negative curvature from the Hessian matrix with principled sequence starting from noise to obtain $O(d)$ escape time. 
Mostly recently, \cite{jin2017escape} improved the escape time complexity to $O(\log^4(d))$ with sphered noise in gradient descent.

By contrast, we show that by extracting curvature through historical stochastic gradient information to guide Langevin dynamics, our proposed ASGLD framework and its specialized versions escape from saddle points with only $O(\log d)$ dependence on dimensionality.
To our knowledge, this is by far the best result for first-order methods. 
Moreover, ASGLD holds two major advantages over existing methods:
i) Unlike \citep{jin2017escape,xu2017first} where a ``stop-at-$\x_{k}$-and-try'' subroutine is needed for saddle point escape, our ASGLD method accomplishes the escape without stopping, which significantly reduces the complexity of the algorithm.
ii) All existing methods \citep{jin2017escape,xu2017first} require full gradient information in identifying escape direction, which implies that they do not work with cases where only stochastic gradients are available.
On the contrary, our ASGLD framework work {\em directly} with stochastic gradients, which are the case for most learning problems in practice.
We summarize the above discussions in in Table \ref{tbl1}, where it can be seen that our ASGLD framework outperforms all known first-order methods and remains competitive compared to significantly more expensive second-order methods.

\begin{table}
	\centering
	\caption{Summary of saddle point escape time complexities for existing methods.}\label{tbl1}
	\begin{tabular}{cllc}
		\hline
		Regime & Algorithm & Time complexity & Oracle\\
		\hline
		\multirow{3}{*}{Second-Order}&\cite{allen2017natasha} & $O(1/\epsilon^{3.25})$ & Hessian, Stochastic gradient\\
		& \cite{nesterov2006cubic} & $O(1/\epsilon^{1.5})$ & Hessian\\
		& \cite{curtis2017trust} & $O(1/\epsilon^{1.5})$ & Hessian\\
		\hline
		Indirect&\cite{agarwal2016finding} & $O(\log d/\epsilon^{7/4})$ & Hessian-vector product\\
		Second-Order& \cite{carmon2016accelerated} & $O(\log d/\epsilon^{7/4})$ &Hessian-vector product\\
		& \cite{carmon2016gradient} & $O(\log d/\epsilon^2)$ &Hessian-vector product\\
		\hline
		\multirow{8}{*}{First-Order}&\cite{ge2015escaping} & $O\left(\mathrm{poly}(d/\epsilon) \right) $ & Stochastic Gradient\\
		&\cite{jin2017escape} & $O\left( \log^4(d)/\epsilon^2 \right) $ &  Gradient\\
		&\cite{levy2016power} & $O\left( d^3\mathrm{poly}(1/\epsilon)\right) $ & Batch Stochastic Gradient\\
		&\cite{xu2017first}      & $\tilde{O}(d/\epsilon^{3.5})$ & Gradient\\
		&	\textbf{Our work} AGLD & $\tilde O(\log d/\epsilon^2)$ & Gradient\\
		&\textbf{Our work} ASG  & $\tilde O(\log d/\epsilon^4)$ & Stochastic Gradient\\
		&\textbf{Our work} ASGLD  & $\tilde O(\log d/\epsilon^4)$ & Stochastic Gradient\\
		&\textbf{Our work} ASGLD-batch  & $\tilde O(\log d/\epsilon^2)$ & Batch Stochastic Gradient\\
		\hline
	\end{tabular}
\end{table}

\section{The adaptive stochastic gradient Langevin dynamics framework}\label{sec::algorithm}

In this section, we first present the most general adaptive stochastic gradient Langevin dynamics (ASGLD) framework, which is followed by two specialized versions, namely the adaptive gradient Langevin dynamics (AGLD) algorithm and the adaptive stochastic gradient (ASG) algorithm.

{\bf Notation: } In this paper, we use boldface to denote matrices/vectors.
We use $\mathrm{Diag}\{\v\}$ to represent the diagonal matrix with $\v$ on its main diagonal.
We let $\mathrm{Tr}\{\A\}$ be the trace of $\A$.
We let $[\A]_{ij}$ represent the entry in the $i$-th row and $j$-th column of $\A$ and let $[\v]_{m}$ represent the $m$-th entry of $\v$.
We let $\I_{N}$ and $\mathbf{O}_{N}$ denote the $N\times N$ identity and all-zero matrices, respectively. 
We will often omit ``$N$'' for brevity if the dimension is clear from the context.
We use $\|\cdot\|$ and $\|\cdot\|_{1}$ to denote $L^{2}$- and $L^{1}$-norms, respectively.
We use ``$\star$" to represent the element-wise product. 
Define $A=O(B)$ if $\| A^{-1}B\|$ converges to some finite constant. Let $A=\tilde{O}(B)$ if $\|A^{-1} B \log^m(B) \|$ converges to some finite constant for some non-negative integer $m$.
We use $\nabla f(\cdot)$ to represent the gradient of $f(\cdot)$ and $\nabla^2 f(\cdot)$ is the Hessian of $f(\cdot)$.
We let $\lambda_{\min}\{\A\}$ and $\lambda_{\max}\{\A\}$ denote the smallest and largest eigenvalues of $\A$, respectively.

\smallskip
Recall that we are interested in solving a non-convex optimization problem in the following form:
\begin{align*} 
\min_{\x \in \mathbb{R}^{d}} f(\x) \triangleq \min_{\x \in \mathbb{R}^{d}} \frac{1}{n} \sum_{i=1}^{n} F(\x,\xi_{i}),  
\end{align*}
where $\x$ is a $d$-dimensional parameter vector and $\{\xi_i\}_{i=1}^n$ are $n$ random samples. 
To solve this problem, we propose a new ASGLD algorithmic framework, which is stated in Algorithm~\ref{ASGLD}.
We will show that the proposed ASGLD framework converges with a constant step size $\eta$, for which the traditional SGLD fails to converge. 
As shown in Algorithm~\ref{ASGLD}, the convergence speed of the algorithm is controlled by parameter $\alpha$, and the escape time from a saddle point is governed by parameter $\beta$. 
The adaptive diagonal matrix series $\{\D_{k}\}_{k=1}^{\infty}$ incrementally acquire the geometry of the objective function contours using historical gradient information. 
Note that since each $\D_k$ is a diagonal matrix, the corresponding inverse computation and storage can be decentralized, which enable distributed algorithm designs.
To see this, we can rewrite (\ref{asgld2}) as: 
\begin{align*}
[\x_{k+1}]_{\ell}=[\x_{k}]_{\ell}-\eta [\D_{k-1}]_{\ell\ell}^{-\alpha/2} [G(\x_{k})]_{\ell}+\sqrt{2\eta}[\D_{k-1}]_{\ell\ell}^{-\beta/2}[\e_{k}]_{\ell},
\end{align*}
separately for $\ell=1,2,\cdots,d$, where $[\x_{k}]_{\ell}$ is the $\ell$-th entry of $\x_{k}$, $[\D_{k-1}]_{\ell\ell}$
is the $\ell$-th diagonal entry in $\D_{k-1}$, $[G(\x_{k})]_{\ell}$ is the $\ell$-th element in $G(\x_{k})$, and $[\e_{k}]_{\ell}$ is the $\ell$-th entry of $\e_k$. 
This illustrates that our proposed ASGLD performs stochastic gradient Langevin dynamics for each dimension in parallel but with different adaptive step sizes.
The major advantage of this adaptive updating is that it gives frequently occurring features lower learning rates and infrequent features higher learning rates, which proves to be more efficient in high-dimensional settings.
The problem of how to balance the convergence speed and the saddle point escape time by tuning parameters $\alpha$ and $\beta$ will be established later in the next section. 

\begin{algorithm}[t!]
	\caption{Adaptive Stochastic Gradient Langevin Dynamics (ASGLD)}\label{ASGLD}
	\begin{algorithmic} 
		\REQUIRE Input step size $\eta>0$,  positive numbers $\sigma^2>0,\alpha, \beta$; initialize $\x_0=\x_1$, and $\D_{-1}=\delta \mathrm{\bm I}_{d\times d}$ with a small positive number $\delta$. 
		\FOR{$k=1,2,\cdots$:} 
		\STATE 1. Sample $\e_k \sim N(0, \sigma^2\mathrm{\bm I}_{d\times d})$. 
		\STATE 2. Compute a stochastic gradient $G(\x_{k})$ that satisfies $\mathbb{E}\left\lbrace G(\x_{k})\mid \x_{k}\right\rbrace=\nabla f(\x_{k})$.
		\STATE 3. Set $\D_{k}=\D_{k-1}+\mathrm{Diag}\left\lbrace G(\x_{k-1})\star G(\x_{k-1})\right\rbrace +\delta \I_{d\times d}$. 
		\STATE 4. Compute the next iterate $\x_{k+1}$ as:
		\begin{eqnarray}
		\x_{k+1}=\x_{k}-\eta \D_{k-1}^{-\alpha/2} G(\x_{k})+\sqrt{2\eta}\D_{k-1}^{-\beta/2}\e_{k}.\label{asgld2}
		\end{eqnarray}
		\ENDFOR
	\end{algorithmic}
\end{algorithm}

Note that our proposed ASGLD is a unifying algorithmic framework in the sense that it covers many first-order methods as special cases.
For example, if $\alpha=0$, $\beta=0$, and $\sigma^2=u^{-1}$, ASGLD reduces to the traditional SGLD. 
In this paper, we pay particular attention to two interesting special cases, namely adaptive gradient Langevin dynamics (AGLD) and adaptive stochastic gradient (ASG).
First, when full gradient $\nabla f(\x_{k})$ is available for $\x_{k}$, $\forall k$, ASGLD becomes AGLD as shown in Algorithm~\ref{AGLD}.
Analyzing the performance of AGLD and comparing it to that of ASGLD will help us understand the price we need to pay and performance loss for using stochastic gradients.

\begin{algorithm}[h!]
	\caption{Adaptive  Gradient Langevin Dynamics (AGLD)}\label{AGLD}
	\begin{algorithmic} 
		\REQUIRE Input step size $\eta>0$,  $\sigma^2>0,\alpha, \beta$; initialize $\x_0=\x_1$, and $\D_{-1}=\delta \I_{d\times d}$ with a small positive number $\delta$. 
		\FOR{$t=1,2,\cdots$:} 
		\STATE 1. Sample $\e_k \sim N(0, \sigma^2 \I_{d\times d})$. 
		\STATE 2. Evaluate the full gradient $\nabla f(\x_{k})$.
		\STATE 3. Set $\D_{k}=\D_{k-1}+\mathrm{Diag}\left\lbrace \nabla f(\x_{k-1}) \star \nabla f(\x_{k-1})\right\rbrace+\delta \I_{d\times d} $. 
		\STATE 4. Compute the next iterate $\x_{k+1}$ as:
		\begin{eqnarray}
		\x_{k+1}=\x_{k}-\eta \D_{k-1}^{-\alpha/2}\nabla f(\x_{k})+\sqrt{2\eta}\D_{k-1}^{-\beta/2}\e_{k}.\label{agld}
		\end{eqnarray}
		\ENDFOR
	\end{algorithmic}
\end{algorithm}

Also, by letting $\beta \uparrow \infty$, we recover the ASG algorithm as illustrated in Algorithm~\ref{ASG}.
\begin{algorithm}
	\caption{Adaptive Stochastic Gradient(ASG)}\label{ASG}
	\begin{algorithmic} 
		\REQUIRE Input step size $\eta>0$,  positive numbers $\sigma^2>0,\alpha$; initialize $\x_0=\x_1$, and $\D_{-1}=\delta \I_{d\times d}$ with a small positive number $\delta$. 
		\FOR{$t=1,2,\cdots$:} 
		\STATE 1. Compute a stochastic gradient $G(\x_{k})$ that satisfies $\mathbb{E}\left\lbrace G(\x_{k})\mid \x_{k}\right\rbrace=\nabla f(\x_{k})$.
		\STATE 2. Set $\D_{k}=\D_{k-1}+\mathrm{Diag}\left\lbrace G(\x_{k-1})\star G(\x_{k-1})\right\rbrace +\delta \I_{d\times d}$. 
		\STATE 3. Compute the next iterate $\x_{k+1}$ as:
		\begin{eqnarray}
		\x_{k+1}=\x_{k}-\eta \D_{k-1}^{-\alpha/2} G(\x_{k}).\label{asg}
		\end{eqnarray}
		\ENDFOR
	\end{algorithmic}
\end{algorithm}

Note that, we can rewrite (\ref{asg}) as:
\begin{eqnarray}
\x_{k+1}=\x_k - \eta \D_{k-1}^{-\alpha/2} \nabla f(\x_k)+\eta \D^{-\alpha/2} \left\lbrace \nabla f(\x_k)-G(\x_{k})\right\rbrace.\nonumber 
\end{eqnarray}
If treating the variability from stochastic gradient $\nabla f(\x_k)-G(\x_{k})$ as  the noise $\e_k$ and letting $\beta=\alpha$,  ASG can also be viewed as a special case of AGLD.  
The advantage of using ASGLD is that we can inject arbitrary noise $\e_k$ and have one addition degree of freedom to manipulate $\beta$, which allows us to exert further control on saddle point escape time.

\section{Main theoretical results}\label{theory}

\subsection{Convergence analysis}

In this section, we first analyze the convergence properties for ASGLD and two specialized versions: AGLD and ASG. 
Then, by proving ASGLD's time complexity for saddle point escape, we establish ASGLD's convergence to a local minimum. 
Thanks to the generality of our ASGLD framework, the theoretical results in this section are applicable to a wide class of existing first-order algorithms.

Before we show main results, we first state several assumptions used in this paper.
\begin{assumption}[Lipschitz Continuity]\label{AA1}
	$f(\x), \nabla f(\x), \nabla^2f(\x)$ are Lipschtiz continous, i.e., there exist three constants $M,L,\rho>0$ such that for any $\x,\y \in \mathbb{R}^{d}$, $\|\nabla f(\x)- \nabla f(\y)\|\leq M \|\x-\y\|$, $|f(\x)-f(\y)|\leq L \|\x-\y\|$,  and $\|\nabla^2 f(\x)-\nabla^2 f(\y)\|\leq \rho \|\x-\y\|$.
\end{assumption}

Note that the Lipschitz continuity assumptions are standard and used in most convergence analysis in non-convex learning (e.g., \cite{zhang2017hitting}, \cite{xu2017global}, \cite{teh2016consistency}, etc.).
The assumption on bounded Hessian is also used in \cite{ge2015escaping}.
The next assumption is also standard for analyzing stochastic gradient algorithms.

\begin{assumption}[Unbiased and Bounded Stochastic Gradient] \label{AA3}
	The stochastic gradients satisfies i) $\mathbb{E}\left\lbrace G(\x)\mid \x\right\rbrace=\nabla f(\x)$ and ii) $\mathrm{Var}\left\lbrace G(\x) \right\rbrace \leq C/B$ for any $\x$, where $C>0$ is a constant and $B>0$ is the batch size.
\end{assumption}
Assumption~\ref{AA3} says that the stochastic gradient is an unbiased estimator of the full gradient and the sampling variance is bounded by a constant uniformly.  
In this paper, we include the standard SGD method as a special case of the mini-batch version of SGD by letting the batch size $B=1$. 
The following definition describes the exact and the $\epsilon$-first-order stationary points.
\begin{definition}
	For a differentiable function $f(\cdot)$,  $\x$ is a first-order stationary point, if $\|\nabla f(\x)\|=0$.  Furthermore, $\x$ is an $\epsilon$-first-order stationary point if $\|\nabla f(\x)\|\leq \epsilon$ for some $\epsilon >0$.
\end{definition}


Now, we state the first key result that lays the foundation for all our subsequent convergence analysis for ASGLD:

\begin{theorem}\label{Theorem1}
	Under Assumptions~\ref{AA1} and \ref{AA3}, if the step size $\eta$ is fixed and satisfies $\eta < 1/M$, then ASGLD converges to an $\epsilon$-first-order stationary point if $I_{1},I_{2},I_{3}\leq \frac{\epsilon^{2}}{3}$, where 
	\begin{align*}
	I_1\!\triangleq\! \frac{  \frac{2}{\eta}\left\lbrace f(\x_0)-\mathbb{E}f(\x_k)\right\rbrace }{\sum\limits_{j=0}^{k-1} \! \mathrm{Tr}\left\lbrace \mathbb{E}\left( \D_{j-1}^{-\alpha/2}\right)\right\rbrace},
	I_2\!\triangleq\!\frac{M\eta C\sum\limits_{j=0}^{k-1} \! B_j^{-1}\mathrm{Tr}\left\lbrace  \mathbb{E}\left( \D_{j-1}^{-\alpha}\right)\right\rbrace}{\sum\limits_{j=0}^{k-1} \! \mathrm{Tr}\left\lbrace \mathbb{E}\left( \D_{j-1}^{-\alpha/2}\right)\right\rbrace}, 
	I_3\!\triangleq\!\frac{2M\sigma^2 \sum\limits_{j=0}^{k-1} \! \mathrm{Tr}\left\lbrace  \mathbb{E}\left( \D_{j-1}^{-\beta}\right) \right\rbrace}{\sum\limits_{j=0}^{k-1} \! \mathrm{Tr}\left\lbrace \mathbb{E}\left( \D_{j-1}^{-\alpha/2}\right)\right\rbrace}.\nonumber
	\end{align*}
\end{theorem}


Roughly speaking, in Theorem~\ref{Theorem1}, $I_1$ describes the convergence of stochastic gradients, $I_2$ describes the convergence of noise from stochastic gradient, and $I_3$ describes the convergence of the injected random noise. Then,  how to choose an appropriate combination of $(\alpha, \beta)$ to make $I_1, I_2$ and $I_3$ be simultaneously small is becomes the key for convergence speed. 
Note that the convergence of ASGLD cannot be obtained for arbitrary choices of $(\alpha, \beta)$.
By specializing ASGLD to AGLD and ASG, the convergence of  AGLD and ASG can be established as follows:

\begin{proposition}\label{proposition2}
	Under Assumptions~\ref{AA1} and \ref{AA3} and if $\eta < 1/M$, AGLD converges to an $\epsilon$-first-order stationary point if $I_1\leq \epsilon^2/2$ and $I_3\leq\epsilon^2/2$.
\end{proposition}

\begin{proposition}\label{proposition3}
	Under Assumptions~\ref{AA1} and \ref{AA3} and if $\eta < 1/M$, ASG converges to an $\epsilon$-first-order stationary point if $I_1\leq \epsilon^2/2$ and $	I_2\leq\epsilon^2/2$.
\end{proposition}

Note that since the AGLD can leverage full gradient information, we have $C=0$.
Hence, $I_2=0$ and not needed in AGLD. 
$I_1$ and $I_3$ are the same as in ASGLD. 
For ASG, since it is a special case with $\beta \uparrow \infty$, we have that $I_3=0$ and hence only $I_1$ and $I_2$ are needed.
The theoretical justification of Theorem~\ref{Theorem1} and Propositions~ \ref{proposition2}--\ref{proposition3} are provided in supplementary.
Next, we establish the convergence rate results under various combinations of $(\alpha,\beta)$:

\begin{theorem}\label{Theorem2}
	Under Assumptions~\ref{AA1} and \ref{AA3}, if $\eta < 1/M$ and the batch size $B_k=B$ is a constant,
	then ASGLD converges to an $\epsilon$-first-order stationary point if $\alpha \in (0,2]$ and $\beta \in (\alpha/2, \infty)$ and with the following convergence rates:
	\vspace{-.07in}
	\begin{list}{\labelitemi}{\leftmargin=1em}
		\itemsep -6pt
		
		\item If $\alpha=2, \beta>1$, ASGLD converges at rate $O\left( \log^{-1}k \right)$;
		\item If $\alpha\in(1,2), \beta \geq 1$, ASGLD converges at rate $\tilde O\left( k^{-\beta+\alpha/2}\right)$; Otherwise, if $\beta < 1$, ASGLD converges at rate $O\left( k^{-1+\alpha/2} \right)$;
		\item If $\alpha=1$ and if $\beta >1$, ASGLD converges at rate $\tilde O(k^{-1/2})$; Otherwise, if $\frac{1}{2} < \beta \leq 1$, ASGLD converges at rate $O(k^{-\beta+1/2})$;
		\item If $\alpha\in(0,1), \beta>\alpha$, ASGLD converges at rate $O\left( k^{-\alpha/2} \right)$; Otherwise, if $\frac{\alpha}{2} < \beta \leq \alpha$, ASGLD converges at rate $O\left( k^{-\beta+\alpha/2}\right)$.
	\end{list}
\end{theorem}

\begin{wrapfigure}{r}{0.4\textwidth}
	\vspace{-20pt}
	\begin{center}
		\includegraphics[width=0.38\textwidth]{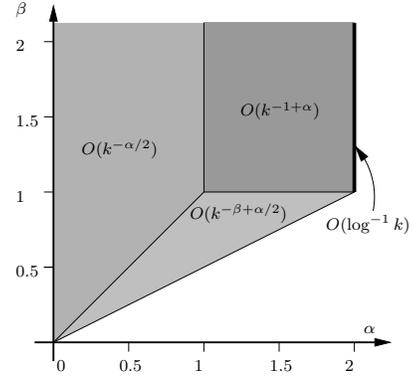}
	\end{center}
	\vspace{-.1in}
	\caption{AGSLD convergence regions.}\label{fig1}
	\vspace{-.1in}
\end{wrapfigure}

Two remarks for Theorem~\ref{Theorem2} are in order:
i) By letting $\alpha\xrightarrow{}0$ and $\beta\xrightarrow{}0$, we would recover the traditional SGLD.
In this case, $I_1 \rightarrow 0$. 
However, $I_2=M\eta C/B>0$ and $I_3=2M\sigma^2>0$. 
Therefore, SGLD does not converge to a stationary point, but only a neighborhood of the stationary point. 
ii) Theorem~\ref{Theorem2} says that large $\alpha$ leads to a quick convergence of $I_2$ but a slow rate of $I_1$. 
Thus, there exists a trade-off between $I_1$ and $I_2$.
Note that $\beta>\alpha/2$ is a sufficient condition to achieve convergence. 
If $\beta$ is large, ASGLD converges to a stationary point quickly,
but the saddle point escape time may increase. 
Thus, the proof of Theorem~\ref{Theorem2} shows that we need to choose $(\alpha,\beta)$ carefully to strike a good  balance between convergence and escaping time.
We summarize the convergence regions of ASGLD in Fig.~\ref{fig1} for easy visualization.


By allowing the batch size to be increasing, we can strengthen the results in Theorem~\ref{Theorem2} as follows:
\begin{theorem}\label{Theorem_batch}
	Under Assumptions~\ref{AA1} and \ref{AA3}, if $\eta < 1/M$, $B_k=O(k^\theta)$ for an arbitrarily small $\theta> 0$, and $\alpha=0$,
	then the convergence of ASGLD is $\tilde O\left( k^{-\min\{1, \theta,\beta\}}\right)$.
	Further, if $\alpha=\beta=1$ and $B_k=O(k)$ (i.e., linear increasing),  the convergence time complexity of ASGLD is $\tilde O(1/\epsilon^{4})$; if $\alpha=0, \beta=1$, the convergence time complexity of ASGLD is $\tilde O(1/\epsilon^{2})$.
\end{theorem}



We note that when $\alpha=\beta=1$, $I_1, I_2, I_3$ share the same convergence rates of $\tilde O\left( k^{-1/2}\right)$, i.e., the algorithm balances stochastic gradient and random noise. 
We refer to this case as the balanced ASGLD. 
The convergence results of AGLD and ASG follows from Theorem~\ref{Theorem2} immediately.

\begin{proposition}\label{proposition6}
	Under the same assumptions as in Theorem~\ref{Theorem2}, AGLD converges to an $\epsilon$-first-order stationary point if $\alpha\in [0,2]$ and $\beta\in(\alpha/2, \infty)$. 
	AGLD achieves the balance between gradient and noise, when $\alpha=0$ and $\beta=1$. The balanced AGLD obtains an $\epsilon$-first-order stationary point with $\tilde O(1/\epsilon^2)$ iterations.
	Moreover, ASG converges to an $\epsilon$-first-order stationary points if $\alpha\in(0,2]$.  ASG obtains the balance for $\alpha=1$ and converges to an $\epsilon$-first-order stationary points with $\tilde O(1/\epsilon^4)$ iterations.
	The specific convergence rates for various combinations of $(\alpha,\beta)$ are as follows:
	
	\vspace{-.07in}
	\begin{list}{\labelitemi}{\leftmargin=1em}
		\itemsep -6pt
		\item If $\alpha=2, \beta>\alpha/2$, AGLD converges at rate $O(1/\log k)$.
		\item If $\alpha \in (0,2)$ and $\beta\geq 1$, AGLD converges at rate $O\left( k^{\alpha/2-1}\right)$. If $\alpha \in (0,2),\beta>\alpha/2$ and $\beta<1$, AGLD converges at rate $O\left(k^{-\beta+\alpha/2} \right)$.
		\item AGLD converges at rate $\tilde O(k^{-1})$ if $\alpha=0$ and $\beta\geq 1$ and $O(k^{-\beta})$ if $\alpha=0$ and $0<\beta<1$.
		
	\end{list}
\end{proposition}



\subsection{Saddle point escape time analysis}

To analyze saddle point escape time, we first formally define the notion of $\gamma$-strict saddle point.
\begin{definition}
	For some given $\gamma >0$, $\x$ is called a $\gamma$-strict saddle point if $\lambda_{\min}\left\lbrace \nabla^2f(\x)\right\rbrace\leq -\gamma <0$.
\end{definition}

To bound high-order moments of stochastic gradient, we assume the following condition.
\begin{assumption} \label{A4}
	The stochastic gradient $G(\x)$ has sub-Gaussian tails, i.e., there exist a positive constant $C'$ and $v$, such that for every $t>0$, $\mathrm{Pr}\left( \|G(\x)\|>t\right)\leq C'\exp(-vt^2)$.
\end{assumption}
The sub-Gaussian assumption is equivalent to assuming that there exists a positive constant $K$ such that $\mathbb{E}\left(\| G(\x)\|^r \right)^{1/r} <K\sqrt{r}$ is bounded uniformly for any $r$-th moment. 

Note that we have already shown that our proposed ASGLD, AGLD and ASG converge to an $\epsilon$-first-order stationary point. 
If the stationary point is a $\gamma$-strict saddle point, we will show that our proposed algorithms can escape from the saddle point with maximum $O(\log d)$ iterations. 
Therefore, our proposed algorithms guarantee converging to a local minimum.
We state this result as follows:
\begin{theorem} \label{Theorem3}
	Suppose that $\x_{k}$ is $\gamma$-strict saddle in some iteration $k$ and $f(\x)$ satisfies Assumptions~\ref{AA1}--\ref{A4}. 
	The proposed ASGLD, AGLD and ASG escape from $\x_{k}$ with at most $T_{\max}=O\left(\left(  a_1\log d+a_2\min(\alpha,\beta)\right) / (\log \gamma + a_3) \right)$ iterations, where $a_1,a_2$ are positive constants and $a_3$ is some sufficiently large constant such that $(\log \gamma + a_3)$ is positive.
\end{theorem}

Three important remarks for Theorem~\ref{Theorem3} are in order:
i) Theorem \ref{Theorem3} says that our proposed ASGLD algorithmic framework escapes from saddle points with at most $T_{\max}$ iterations. 
Note that $T_{\max}$ only grows at rate $\log d$, which is almost free of dimension. 
This happens because our proposed ASGLD uses adaptive matrix $\D_k$, which scales the step size $\eta$ in different dimensions to leverage the geometry information of objective contour space. 
This significantly helps the algorithm escape from saddle points along multiple coordinates {\em simultaneously}.
ii) The negative eigenvalue $\gamma$ also affects the efficiency of escaping: the smaller $\gamma$, the larger $T_{\max}$ (i.e., harder to escape).  
iii) The escape time depends on the minimum of $(\alpha, \beta)$, which suggests a trade-off between convergence and escape time. 
One one hand, a small $\beta$ implies the small escape time, but a slow convergence rate, due to $I_3$ in Theorem \ref{Theorem1}. On the other hand, using a large $\beta$ leads to a quick convergence of $I_3$, but a large escape time. $\alpha=\beta=1$ balances the convergence rate and escape time.

Combining Theorems~\ref{Theorem2} and \ref{Theorem3}, we establish our final theoretical results as follows.

\begin{theorem}\label{theorem4}
	Suppose that Assumptions~\ref{AA1}--\ref{A4} are satisfied, all saddle points of $f(\x)$ are $\gamma$-strict, and $\eta < 1/M$, then the balanced ASGLD and ASG converge to a local minimum with at most $\tilde O( \frac{\log d}{\epsilon^4}) $; 
	and the balanced AGLD converges to a local minimum with at most $\tilde O(\frac{\log d}{\epsilon^2}) $;
	and ASGLD with increasing batch size converges to a local minimum with at most $\tilde O( \frac{\log d}{\epsilon^2}) $ if $B_k=O(k)$.
\end{theorem}
Theorem~\ref{theorem4} indicates that AGLD converges quickly to local minimum if full gradient information can be obtained easily. 
While having a slower convergence rate, ASGLD and ASG have lower computation complexity per iteration by using stochastic gradient, and this is particularly true when the sample space size $n$ is large. 
ASGLD with increasing batch size can also achieve the best convergence rate as long as $B$ is increasing linearly, regardless how small the slope is.

\section{Conclusion} \label{sec::conclusion}
In this paper, we developed a new adaptive stochastic gradient Langevin dynamics framework (ASGLD).
We also studied its two specialized versions, adaptive gradient Langevin dynamics (AGLD) and adaptive stochastic gradient (ASG) algorithms. 
We established their convergence rates and saddle point escape times. 
The parameter combinations that balance between convergence rate and escape time are carefully investigated. 
We conducted numerical studies (presented in supplementary material) to verify the performance of the proposed algorithms.
Our methods significantly generalized the traditional stochastic gradient Langevin dynamics
Our results showed that the proposed algorithms escape from saddle points with at most $O(\log d)$ iterations, which is nearly free of dimension. 
Our proposed algorithms consider smooth objective functions. 
The extension to non-smooth objective functions will be considered in our future work.

\clearpage

\bibliographystyle{chicago}
\bibliography{ref}

\clearpage
\appendix

In this document,  we provide proof details for all theoretical results in this paper.

\section{Preliminaries}
Before we show the justifications, we first prepare some lemmas, that will be useful for subsequent analysis.

\begin{lemma}\label{lemma1}
	Let $\D_k$ be the diagnostic matrix as computed in ASGLD, AGLD and ASG, respectively. It then follows that
	\begin{eqnarray}
	\mathbb{E}\left[ \D_{k-1}^{-\alpha}\right] =O\left( k^{-\alpha}\right), \nonumber
	\end{eqnarray}
	Similarly, 
	\begin{eqnarray}
	\mathbb{E}\left[ \D_{k-1}^{-\alpha/2}\right] =O\left( k^{-\alpha/2}\right), \nonumber\\
	\mathbb{E}\left[ \D_{k-1}^{-\beta}\right] =O\left( k^{-\beta}\right) \nonumber.
	\end{eqnarray}
\end{lemma}

\begin{proof}
	For ASGLD and ASG, the definition of $\D_{k-1}$ is 
	\begin{eqnarray}
	\D_{k-1}=\sum_{j=0}^{k-1} \mathrm{Diag}\left\lbrace G(\x_j)\star G(\x_j)\right\rbrace +\delta\mathrm{\bm I}.\nonumber
	\end{eqnarray}
	Let $[\D_{k-1}]_{\ell,\ell}$ be $\ell$-th diagnostic entry in $\D_{k-1}$. Then, we have  $[\D_{k-1}]_{\ell,\ell}>\delta k$. Note that, the $\ell$-th diagnostic entry in $\mathbb{E}\left( \D_{k-1}^{-\alpha}\right) $ is $\mathbb{E}\left( [\D_{k-1}]_{\ell,\ell}^{-\alpha}\right) $. Thus,\begin{eqnarray}
	\mathbb{E}\left( [\D_{k-1}]_{\ell,\ell}^{-\alpha}\right) <\delta^{-\alpha} k^{-\alpha}=O(k^{-\alpha}).\label{eq1}
	\end{eqnarray}
	
	Since $h(u)=u^{-\alpha}$ is a convex function, 	by Jensen's inequality, we have 
	\begin{eqnarray}
	\mathbb{E}\left( [\D_{k-1}]_{\ell,\ell}^{-\alpha}\right) \geq \left( \mathbb{E}[\D_{k-1}]_{\ell,\ell}\right)^{-\alpha},\nonumber
	\end{eqnarray}
	Moreover, 
	\begin{eqnarray}
	\mathbb{E}\left( [\D_{k-1}]_{\ell,\ell}\right)=\mathbb{E}\left\lbrace \sum_{j=0}^{k-1}[ G^2(\x_j)]_{\ell}+\delta\right\rbrace \leq k\delta +\sum_j[\nabla f^2(X_j)]_{\ell,\ell}+kC= O(k)\nonumber
	\end{eqnarray}
	using \textbf{Assumption 3}. Therefore, 
	\begin{eqnarray}
	\mathbb{E}\left( [\D_{k-1}]_{\ell,\ell}^{-\alpha}\right) \geq O\left( k^{-\alpha}\right)\label{eq2} .
	\end{eqnarray}
	Combing (\ref{eq1}) and (\ref{eq2}), we can conclude that $	\mathbb{E}\left( \D_{k-1}^{-\alpha}\right) =O\left( k^{-\alpha}\right)$.
\end{proof}

For AGLD, the only difference is that $C=0$, which does not effect our final conclusion. Similar procedures can be used for $-\alpha/2$ and $\beta$.  This completes the proof for Lemma \ref{lemma1}.

\begin{lemma}\label{lemma2}
	Let $\D_{k-1}$ be the diagnostic matrix as computed in  ASGLD, AGLD and ASG, respectively. It then follows that:
	\begin{eqnarray}
	\sum_{j=0}^{k-1} \mathrm{Trace}\left\lbrace \mathbb{E}\left( \D_{j-1}^{-\phi}\right) \right\rbrace =\left\lbrace 
	\begin{array}{ll}
	O(dk) & \text{if $\phi=0$}\\
	O(d\log k) & \text{if $\phi=1$}\\
	O(d k^{1-\phi}) & \text{otherwise}.
	\end{array}\right.
	\end{eqnarray}
\end{lemma}

\begin{proof}
	For case $\phi=0$, it is trivial. For $\phi>0$,
	using Lemma \ref{lemma1}, we have:
	\begin{eqnarray}
	\mathbb{E}\left( [\D_{k-1}]_{\ell,\ell}^{-\phi}\right)=O(k^{-\phi}),\quad \text{for} \quad \ell=1,2,\cdots, n,\nonumber
	\end{eqnarray}	
	which implies
	\begin{eqnarray}
	\mathrm{Trace}\left\lbrace \mathbb{E}\left( \D_{k-1}^{-\phi}\right) \right\rbrace =O(d k^{-\phi}).\label{eq3}
	\end{eqnarray}
	This completes the proof by taking summation for (\ref{eq3}) from 0 to $k-1$.
\end{proof}

\section{Proof  for Theorem 4.1}

Using Lipschitz and smoothness assumptions, we have 
\begin{eqnarray}
f(\x_{k+1})\leq f(\x_{k})+\nabla f^T(\x_{k})(\x_{k+1}-\x_{k})+\frac{M}{2}\|\x_{k+1}-\x_{k}\|^2.\label{leq}
\end{eqnarray}
By our proposed ASGLD algorithm, the iteration is updated as
\begin{eqnarray}
\x_{k+1}=\x_{k}-\eta \D_{k-1}^{-\alpha/2} G(\x_{k})+\sqrt{2\eta}\D_{k-1}^{-\beta/2}\e_{k}.\nonumber
\end{eqnarray}
Thus, taking the expectation respect to the stochastic gradient and the noise, we can obtain that 
\begin{eqnarray}
\mathbb{E}\left\lbrace \x_{k+1}-\x_{k}\mid \x_{k}\right\rbrace=-\eta \mathbb{E}\left( \D_{k-1}^{-\alpha/2}\right)  \mathbb{E}\left\lbrace G(\x_{k})\mid \x_{k}\right\rbrace+\sqrt{2\eta} \mathbb{E}\left( \D_{k-1}^{-\beta/2}\right) \mathbb{E}\left(\e_k\right) .\label{ep}
\end{eqnarray}
Since $\mathbb{E}\left(\e_k\right)=\bm 0$ and $\mathbb{E}\left\lbrace G(\x_{k})\mid \x_{k}\right\rbrace=\nabla f(\x_k)$, equation (\ref{ep}) leads to 
\begin{eqnarray}
\mathbb{E}\left\lbrace \x_{k+1}-\x_{k}\mid \x_{k}\right\rbrace=-\eta \mathbb{E}\left( \D_{k-1}^{-\alpha/2}\right) \nabla f(\x_k).\label{claim 1}
\end{eqnarray}
Moreover, the quadratic term in (\ref{leq}) can be rewritten as
\begin{eqnarray}
\|\x_{k+1}-\x_{k}\|^2=\eta^2G^T(\x_{k})\D_{k-1}^{-\alpha}G(\x_k)+2\eta \e_k^T \D_{k-1}^{-\beta}\e_k-2\eta\sqrt{2\eta}G^T(\x_{k})\D_{k-1}^{-(\alpha+\beta)/2}\e_k.\label{quad}
\end{eqnarray}

Now, we invoke a fact for the expectation of a quadratic function \citep{mathai1992quadratic2}. 
\begin{fact}\label{fact}
	For a vector random variable $\x$, which satisfies $\mathbb{E}(\x)=\bmu$ and $\mathrm{Var}(\x)=\Sigma$ and a symmetric matrix $\Lambda$, we have 
	\begin{eqnarray}
	\mathbb{E}\left( \x^T\Lambda \x\right)=\mathrm{Trace}(\Lambda\Sigma)+\bmu^T\Lambda \bmu.\nonumber 
	\end{eqnarray}
\end{fact}
Using this fact and taking the expectation of (\ref{quad}), we can obtain that 
\begin{eqnarray}
&\mathbb{E}\left( \|\x_{k+1}-\x_{k}\|^2\mid \x_{k}\right) &=\eta^2\mathrm{Trace}\left[ \mathbb{E}\left( \D_{k-1}^{-\alpha}\right) \mathrm{Var}\left\lbrace G(\x_{k}) \right\rbrace \right] +\eta^2\nabla f^T(\x_{k})\mathbb{E}\left( \D_{k-1}^{-\alpha}\right) \nabla f(\x_{k})\nonumber\\
&&+2\eta\mathrm{Trace}\left\lbrace  \mathbb{E}\left( \D_{k-1}^{-\beta}\right) \sigma^2\right\rbrace \nonumber\\
&&\leq \eta^2 C B_k^{-1}\mathrm{Trace}\left\lbrace  \mathbb{E}\left( \D_{k-1}^{-\alpha}\right)\right\rbrace+2\eta\sigma^2 \mathrm{Trace}\left\lbrace  \mathbb{E}\left( \D_{k-1}^{-\beta}\right) \right\rbrace\nonumber\\
&&+\eta^2\nabla f^T(\x_{k})\mathbb{E}\left( \D_{k-1}^{-\alpha}\right) \nabla f(\x_{k}),\label{quad2}
\end{eqnarray}
by \textbf{Assumption 2}.

Plugging (\ref{claim 1}) and (\ref{quad2}) into (\ref{leq}), we have
\begin{eqnarray}
&\hspace{-.2in}\mathbb{E}\left\lbrace f(\x_{k+1})\mid \x_{k}\right\rbrace -f(\x_{k})&\leq -\eta \nabla f^T(\x_{k})\mathbb{E}\left( \D_{k-1}^{-\alpha/2}\right)\nabla f(\x_{k}) +\frac{M}{2}\eta^2 f^T(\x_{k})\mathbb{E}\left( \D_{k-1}^{-\alpha}\right) \nabla f(\x_{k})\nonumber\\
&&+\frac{M}{2}\eta^2CB_k^{-1} \mathrm{Trace}\left\lbrace  \mathbb{E}\left( \D_{k-1}^{-\alpha}\right)\right\rbrace+M\eta\sigma^2\mathrm{Trace}\left\lbrace  \mathbb{E}\left( \D_{k-1}^{-\beta}\right) \right\rbrace.\nonumber
\end{eqnarray}

For any diagonal matrix $\D$ and vector  $\bm v$, we have $\bm v^T\D \bm v\leq \mathrm{Trace}(\D)\bm v^T\bm v$. Therefore, we have:
\begin{multline}
\mathbb{E}\left\lbrace f(\x_{k+1})\mid \x_{k}\right\rbrace -f(\x_{k}) \leq-\eta\left[ \mathrm{Trace}\left\lbrace \mathbb{E}\left( \D_{k-1}^{-\alpha/2}\right)\right\rbrace-\frac{M}{2}\eta \mathrm{Trace}\left\lbrace \mathbb{E}\left( \D_{k-1}^{-\alpha}\right)  \right\rbrace  \right] \times \\
\nabla f^T(\x_{k})\nabla f(\x_{k})+\frac{M}{2}\eta^2CB_k^{-1} \mathrm{Trace}\left\lbrace  \mathbb{E}\left( \D_{k-1}^{-\alpha}\right)\right\rbrace+M\eta\sigma^2\mathrm{Trace}\left\lbrace  \mathbb{E}\left( \D_{k-1}^{-\beta}\right) \right\rbrace.\nonumber
\end{multline}

Let all diagonal entries in $\D_{-1}$ be greater than or equal to 1.  We have $\D_{k-1}-\mathrm{\bm I}$ is positive definite. Then, 
\begin{eqnarray}
\mathrm{Trace}\left\lbrace \mathbb{E}\left( \D_{k-1}^{-\alpha/2}\right)\right\rbrace\geq  \mathrm{Trace}\left\lbrace \mathbb{E}\left( \D_{k-1}^{-\alpha}\right)\right\rbrace.\nonumber
\end{eqnarray}
Since $1-M\eta/2>1/2$, we have 
\begin{eqnarray}
&\mathbb{E}\left\lbrace f(\x_{k+1})\mid \x_{k}\right\rbrace -f(\x_{k})&\leq-\frac{\eta}{2} \mathrm{Trace}\left\lbrace \mathbb{E}\left( \D_{k-1}^{-\alpha/2}\right)\right\rbrace \nabla f^T(\x_{k})\nabla f(\x_{k})\nonumber\\
&&+\frac{M}{2}\eta^2CB_k^{-1} \mathrm{Trace}\left\lbrace  \mathbb{E}\left( \D_{k-1}^{-\alpha}\right)\right\rbrace+M\eta\sigma^2\mathrm{Trace}\left\lbrace  \mathbb{E}\left( \D_{k-1}^{-\beta}\right) \right\rbrace,\nonumber
\end{eqnarray}
which further implies that
\begin{eqnarray}
&&\mathbb{E}\left\lbrace f(\x_{k+1})\right\rbrace -\mathbb{E}\left\lbrace f(\x_{k})\right\rbrace \leq-\frac{\eta}{2} \mathrm{Trace}\left\lbrace \mathbb{E}\left( \D_{k-1}^{-\alpha/2}\right)\right\rbrace \mathbb{E}\left\lbrace  \nabla f^T(\x_{k})\nabla f(\x_{k})\right\rbrace \nonumber\\
&&+\frac{M}{2}\eta^2C{B_k}^{-1} \mathrm{Trace}\left\lbrace  \mathbb{E}\left( \D_{k-1}^{-\alpha}\right)\right\rbrace+M\eta\sigma^2\mathrm{Trace}\left\lbrace  \mathbb{E}\left( \D_{k-1}^{-\beta}\right) \right\rbrace.\label{edf}
\end{eqnarray}

Summing (\ref{edf}) from 0 to $t$, we obtain that 
\begin{eqnarray}
&\mathbb{E}\left\lbrace f(\x_{k})\right\rbrace -f(\x_0)&\leq -\frac{\eta}{2}\sum_{j=0}^{k-1}\left[ \mathrm{Trace}\left\lbrace \mathbb{E}\left( \D_{j-1}^{-\alpha/2}\right)\right\rbrace \mathbb{E}\left\lbrace  \nabla f^T(\x_{j})\nabla f(\x_{j})\right\rbrace\right] \nonumber\\
&&+\frac{M\eta^2C}{2}\sum_{j=0}^{k-1}B_j^{-1}\mathrm{Trace}\left\lbrace  \mathbb{E}\left( \D_{j-1}^{-\alpha}\right)\right\rbrace +M\eta\sigma^2 \sum_{j=0}^{k-1} \mathrm{Trace}\left\lbrace  \mathbb{E}\left( \D_{j-1}^{-\beta}\right) \right\rbrace.\nonumber
\end{eqnarray}
Then, we can show that 
\begin{eqnarray}
&&\frac{\eta}{2}\sum_{j=0}^{k-1}\left[ \mathrm{Trace}\left\lbrace \mathbb{E}\left( \D_{j-1}^{-\alpha/2}\right)\right\rbrace \mathbb{E}\left\lbrace  \nabla f^T(\x_{j})\nabla f(\x_{j})\right\rbrace\right]\nonumber\\
&&\leq f(\x_0)-\mathbb{E}f(\x_k)+\frac{M\eta^2C}{2}\sum_{j=0}^{k-1}B_j^{-1}\mathrm{Trace}\left\lbrace  \mathbb{E}\left( \D_{j-1}^{-\alpha}\right)\right\rbrace +M\eta\sigma^2 \sum_{j=0}^{k-1} \mathrm{Trace}\left\lbrace  \mathbb{E}\left( \D_{j-1}^{-\beta}\right) \right\rbrace\nonumber,
\end{eqnarray}
which is equivalent to
\begin{eqnarray}
&&\frac{\sum_{j=0}^{k-1}\left[ \mathrm{Trace}\left\lbrace \mathbb{E}\left( \D_{j-1}^{-\alpha/2}\right)\right\rbrace \mathbb{E}\left\lbrace  \nabla f^T(\x_{j})\nabla f(\x_{j})\right\rbrace\right]}{\sum_{j=0}^{k-1} \mathrm{Trace}\left\lbrace \mathbb{E}\left( \D_{j-1}^{-\alpha/2}\right)\right\rbrace}\nonumber\\
&&\leq \frac{  \frac{2}{\eta}\left\lbrace f(\x_0)-\mathbb{E}f(\x_k)\right\rbrace }{\sum_{j=0}^{k-1} \mathrm{Trace}\left\lbrace \mathbb{E}\left( \D_{j-1}^{-\alpha/2}\right)\right\rbrace}\nonumber\\
&&+\frac{M\eta C\sum_{j=0}^{k-1}B_j^{-1}\mathrm{Trace}\left\lbrace  \mathbb{E}\left( \D_{j-1}^{-\alpha}\right)\right\rbrace}{\sum_{j=0}^{k-1} \mathrm{Trace}\left\lbrace \mathbb{E}\left( \D_{j-1}^{-\alpha/2}\right)\right\rbrace}\nonumber\\
&&+ \frac{2M\sigma^2 \sum_{j=0}^{k-1} \mathrm{Trace}\left\lbrace  \mathbb{E}\left( \D_{j-1}^{-\beta}\right) \right\rbrace}{\sum_{j=0}^{k-1}\mathrm{Trace}\left\lbrace \mathbb{E}\left( \D_{j-1}^{-\alpha/2}\right)\right\rbrace}\nonumber\\
&&=I_1+I_2+I_3.\nonumber
\end{eqnarray}

Note that $$\min_{j}\mathbb{E}\left\lbrace \nabla f^T(\x_{j}) \nabla f(\x_j) \right\rbrace \leq \frac{\sum_{j=0}^{k-1} a_j\mathbb{E}\left\lbrace \nabla f^T(\x_{j}) \nabla f(\x_j) \right\rbrace}{\sum_{j=0}^{k-1} a_j} $$
for any $a_j>0$.

Let $a_j=\eta/2 \mathrm{Trace}\left\lbrace \mathbb{E}\left( \D_{j-1}^{-\alpha/2}\right)\right\rbrace$. We can obtain that 
\begin{eqnarray}
\min_{j}\mathbb{E}\left\lbrace \nabla f^T(\x_{j}) \nabla f(\x_j) \right\rbrace\leq I_1+I_2+I_3.\nonumber
\end{eqnarray}

Therefore, if $(I_1, I_2, I_3)\leq \epsilon^2/3$, we can conclude that $\min_{j}\mathbb{E}\left\lbrace \nabla f^T(\x_{j}) \nabla f(\x_j) \right\rbrace\leq \epsilon^2$, which completes the proof.

\section{Proof for Theorem 4.4}

Using Lemma \ref{lemma2}, we have:
\begin{eqnarray}
\sum_{j=0}^{k-1}\mathrm{Trace}\left\lbrace \mathbb{E}\left( \D_{j-1}^{-\alpha/2}\right) \right\rbrace =\left\lbrace 
\begin{array}{lr}
O(dk) & \text{if $\alpha=0$}\\
O(d\log k) & \text{if $\alpha=2$}\\
O(dk^{1-\alpha/2}) & 0<\alpha<2.
\end{array}\right.\label{trace1}
\end{eqnarray}
Similarly, we can show that:
\begin{eqnarray}
\sum_{j=0}^{k-1}\mathrm{Trace}\left\lbrace \mathbb{E}\left( \D_{j-1}^{-\alpha}\right) \right\rbrace =\left\lbrace 
\begin{array}{lr}
O(dk) & \text{if $\alpha=0$}\\
O(d\log k) & \text{if $\alpha=1$}\\
O(dk^{1-\alpha}) & 0<\alpha\leq 2, \alpha\neq 1,
\end{array}\right.\label{trace2}
\end{eqnarray}
and
\begin{eqnarray}
\sum_{j=0}^{k-1}\mathrm{Trace}\left\lbrace \mathbb{E}\left( \D_{j-1}^{-\beta}\right) \right\rbrace =\left\lbrace 
\begin{array}{lr}
O(dk) & \text{if $\beta=0$}\\
O(d\log k) & \text{if $\beta=1$}\\
O(dk^{1-\beta}) &  \beta\notin \{0,1\}.
\end{array}\right.\label{trace3}
\end{eqnarray}

Using (\ref{trace1}), we have:
\begin{eqnarray}
I_1=\left\lbrace 
\begin{array}{lr}
O((dk)^{-1}) & \text{if $\alpha=0$}\\
O((d\log k)^{-1}) & \text{if $\alpha=2$}\\
O(d^{-1}k^{\alpha/2-1}) & 0<\alpha<2.
\end{array}\right.\nonumber
\end{eqnarray}
Using (\ref{trace1}) and (\ref{trace2}), we can show:
\begin{eqnarray}
I_2=\left\lbrace 
\begin{array}{lr}
O(1) & \text{if $\alpha=0$}\\
O(\log(k)/\sqrt{k}) & \text{if $\alpha=1$}\\
O((k\log k)^{-1}) & \text{if $\alpha=2$}\\
O(k^{-\alpha/2}) & 0<\alpha<2, \alpha \neq 1.
\end{array}\right.\nonumber
\end{eqnarray}
Using (\ref{trace1}) and (\ref{trace3}), we have:
\begin{eqnarray}
I_3=\left\lbrace 
\begin{array}{lr}
O(1)     & \text{if $\beta=0, \alpha=0$}\\
O(k/\log k) & \text{if $\beta=0, \alpha=2$}\\
O(k^{\alpha/2})& \text{if $\beta=0, 0<\alpha<2$}\\
O(k^{-1}\log k) & \text{if $\beta=1,\alpha=0$}\\
O(1) & \text{if $\beta=1, \alpha=2$}\\
O(k^{\alpha/2-1}\log k) & \text{if $\beta=1, 0<\alpha<2$}\\
O(k^{-\beta}) & \text{$\beta \notin\{0,1\}, \alpha=0$}\\
O(k^{1-\beta}/\log k) & \text{$\beta \notin\{0,1\}, \alpha=2$}\\
O(k^{-\beta+\alpha/2}) & \text{$\beta \notin\{0,1\}, 0<\alpha<2$}.
\end{array}\right.\nonumber
\end{eqnarray}

Therefore, if $\alpha \in (0, 2]$ and $\beta\in (\alpha/2, \infty)$, we have:
\begin{eqnarray}
I_1\xrightarrow{}0;\nonumber\\
I_2\xrightarrow{}0;\nonumber\\
I_3\xrightarrow{}0.\nonumber
\end{eqnarray}

The specific convergence rate for ASGLD can be described as follows.
\begin{itemize}
	\item [1.] If $\alpha=2, \beta>1$, we can show that $I_1=O\left( \log^{-1}k \right) , I_2=O\left( k^{-1} \log^{-1}k \right)$ and $I_3=O\left( k^{1-\beta} \log^{-1}k\right) $. Therefore, the convergence rate of ASGLD is determined by $I_1$, which leads to $O\left( \log^{-1}k \right) $.
	\item[2.] If $\alpha\in(1,2), \beta>\alpha/2$, we can show that $I_1=O\left( k^{-1+\alpha/2} \right)$, $I_2=O\left(k^{-\alpha/2} \right)$, and $I_3=\tilde O\left( k^{-\beta+\alpha/2}\right) $. If $\beta\geq 1$, the convergence rate of ASGLD is determined by $I_3$, which is  $\tilde O\left( k^{-\beta+\alpha/2}\right) $.  For $\beta< 1$, the convergence rate is $O\left( k^{-1+\alpha/2} \right)$.
	\item[3.] If $\alpha=1, \beta>1/2$, we can obtain that $I_1=O(k^{-1/2})$, $I_2=O(k^{-1/2}\log k)$ and $I_3=\tilde O(k^{-\beta+\alpha/2})$. Therefore, if $\beta>1$, the convergence rate is $\tilde O(k^{-1/2})$. Otherwise,
	the convergence rate is determined by $I_3=O(k^{-\beta+1/2})$.
	\item[4.] If $\alpha\in(0,1), \beta>\alpha/2$,  we can show that $I_1=O\left( k^{-1+\alpha/2} \right)$, $I_2=O\left(k^{-\alpha/2} \right)$, and $I_3=O\left( k^{-\beta+\alpha/2}\right) $. Therefore, the convergence rate is $I_2=O\left( k^{-\alpha/2} \right) $ if $\beta>\alpha$. Otherwise, the convergence rate is $I_3=O\left( k^{-\beta+\alpha/2}\right) $.
\end{itemize}
This completes the proof for Theorem 4.4.

\section{Proof for Lemma 4.2}

From the proposed AGDL, we have:
\begin{eqnarray}
\x_{k+1}=\x_{k}-\eta \D_{k-1}^{-\alpha/2}\nabla f(\x_{k})+\sqrt{2\eta}\D_{k-1}^{-\beta/2}\e_k,\nonumber
\end{eqnarray}
which implies:
\begin{eqnarray}
\mathbb{E}\left(\x_{k+1}-\x_{k}\mid \x_{k} \right)=-\eta D_{k-1}^{-\alpha/2}\nabla f(\x_{k}).\label{A1} 
\end{eqnarray}
Moreover,  using Fact \ref{fact}, we have:
\begin{eqnarray}
\mathbb{E}\left(\|\x_{k+1}-\x_{k}\|^2\mid \x_{k} \right)=\eta^2 \nabla f^T(\x_{k})\D_{k-1}^{-\alpha} \nabla f(\x_{k})+2\eta\sigma^2 \mathrm{Trace}\left( \D_{k-1}^{-\beta}\right) .\label{A2}
\end{eqnarray}

Plugging (\ref{A1}) and (\ref{A2}) into (\ref{leq}), we obtain:
\begin{eqnarray}
&E\left\lbrace f(\x_{k+1})\mid \x_{k} \right\rbrace -f(\x_{k})\leq& -\eta \nabla f^T(\x_{k})\D_{k-1}^{-\alpha/2}\nabla f(\x_{k})\nonumber\\
&&+\eta^2M/2 \nabla f^T(\x_{k})\D_{k-1}^{-\alpha} \nabla f(\x_{k})+M\eta\sigma^2 \mathrm{Trace}\left( \D_{k-1}^{-\beta}\right)\nonumber\\
&&\leq -\eta/2\nabla f^T(\x_{k}) \nabla f(\x_{k}) \mathrm{Trace}(\D_{k-1}^{-\alpha/2})+M\eta\sigma^2\mathrm{Trace}(\D_{k-1}^{-\beta}).\nonumber
\end{eqnarray}
Taking the marginal expectation, we have 
\begin{multline}
\mathbb{E}\left\lbrace f(\x_{k+1})\right\rbrace-\mathbb{E} \left\lbrace f(\x_{k})\right\rbrace \leq -\eta/2 \mathbb{E}\left\lbrace \nabla f^T(\x_{k}) \nabla f(\x_{k})\right\rbrace \mathbb{E}\left\lbrace \mathrm{Trace}(\D_{k-1}^{-\alpha/2})\right\rbrace \\
+M\eta\sigma^2 \mathbb{E}\left\lbrace \mathrm{Trace}(\D_{k-1}^{-\beta})\right\rbrace.\label{edf2}
\end{multline}
Summing (\ref{edf2}) from 0 to $k-1$,  we can obtain that:
\begin{eqnarray}
&&\eta/2 \sum_{j=0}^{k-1} \mathbb{E}\left\lbrace \nabla f^T(\x_{j}) \nabla f(\x_{j})\right\rbrace \mathbb{E}\left\lbrace \mathrm{Trace}(\D_{j-1}^{-\alpha/2})\right\rbrace\nonumber\\
&&\leq f(\x_0)-\mathbb{E}\left\lbrace f(\x_{k})\right\rbrace+M\eta\sigma^2 \sum_{j=0}^{k-1} \mathbb{E}\left\lbrace \mathrm{Trace}(\D_{j-1}^{-\beta})\right\rbrace.\label{A3}
\end{eqnarray}
Using (\ref{A3}) and similar technique in proof of Theorem 4.1, we have 
\begin{eqnarray}
&\min_{j} \mathbb{E}\left\lbrace \nabla f^T(\x_{j}) \nabla f(\x_{j})\right\rbrace\leq &\frac{ f(\x_0)-\mathbb{E}\left\lbrace f(\x_{k})\right\rbrace}{\eta/2 \sum_{j=0}^{k-1} \mathbb{E}\left\lbrace \mathrm{Trace}(\D_{j-1}^{-\alpha/2})\right\rbrace}+\frac{2M\sigma^2\sum_{j=0}^{k-1} \mathbb{E}\left\lbrace \mathrm{Trace}(\D_{j-1}^{-\beta})\right\rbrace}{ \sum_{j=0}^{k-1} \mathbb{E}\left\lbrace \mathrm{Trace}(\D_{j-1}^{-\alpha/2})\right\rbrace}\nonumber\\
&&=:I_1+I_3\nonumber.
\end{eqnarray}
Thus,  $\min_{j} \mathbb{E}\left\lbrace \nabla f^T(\x_{j}) \nabla f(\x_{j})\right\rbrace\leq \epsilon^2$ if $I_1, I_3\leq \epsilon^2/2$.

The convergence rate for AGLD is shown as follows.
\begin{itemize}
	\item [1.] If $\alpha=2, \beta>\alpha/2$, the convergence rate is $O(1/\log k)$.
	\item[2.] If $\alpha \in (0,2)$ and $\beta\geq 1$, the convergence rate is $O\left( k^{\alpha/2-1}\right)$. If $\alpha \in (0,2),\beta>\alpha/2$ and $\beta<1$, the rate is $O\left(k^{-\beta+\alpha/2} \right)$.
	\item[3.] If $\alpha=0, \beta\geq 1$, the convergence rate is $\tilde O(k^{-1})$. If $\alpha=0, 0<\beta<1$, the convergence rate is $O(k^{-\beta})$.
\end{itemize}

This completes the proof for Lemma 4.2.

\section{Proof for Lemma 4.3}
For ASG, we can rewrite the algorithm as:
\begin{eqnarray}
&\x_{k+1}&=\x_{k}-\eta \D_{k-1}^{-\alpha/2} G(\x_{k})\nonumber\\
&&=\x_{k}-\eta \D_{k-1}^{-\alpha/2} \nabla f(\x_{k}) +\eta \D_{k-1}^{-\alpha/2}\left\lbrace \nabla f(\x_{k})-G(\x_{k})\right\rbrace \nonumber\\
&&=\x_{k}-\eta \D_{k-1}^{-\alpha/2} \nabla f(\x_{k}) +\eta \D_{k-1}^{-\alpha/2}\bm {\zeta}_t,\nonumber
\end{eqnarray}
where $\bm {\zeta}_t=\nabla f(\x_{k})-G(\x_{k})$ and $\mathbb{E}\left( \bm {\zeta}_t\mid \x_{k}\right) =0, \mathrm{Var}\left( \bm {\zeta}_t\mid \x_{k}\right)\leq C/B_k$.
Following the same token in proof for Theorem 4.1, we can conclude that:
\begin{eqnarray}
\min_{j} \mathbb{E}\left\lbrace \nabla f^T(\x_j) f(\x_j)\right\rbrace\leq I_1+I_2.\nonumber 
\end{eqnarray}
Moreover, the convergence rate is $O\left( 1/\log k\right) $, if $\alpha=2$. For $1\leq \alpha<2$, the rate is $O\left( k^{-1+\alpha/2}\right)$. For $0<\alpha<1$, the rate is $O\left( k^{-\alpha/2}\right) $.
This completes the proof for Lemma 4.3.

\section{Proof for Theorem 4.7}
\subsection{Proof for ASGLD and ASG}
Since $\beta\xrightarrow{}\infty$, ASGLD degenerates to ASG. Hence, it is sufficient to show the proof for ASGLD.
Assume that $\x_{k_0}$ is a $\gamma$-strict saddle point, which satisfies $\lambda_{\min}\left\lbrace \nabla^2 f(\x_{k_0}) \right\rbrace\leq -\gamma $ and $\nabla f^T(\x_{k_0})\nabla f(\x_{k_0})\leq \epsilon^2$.

The analysis can be outlined as three steps. 
\begin{itemize}
	\item []\textbf{Step 1}: We show the escaping time for a quadratic function.
	\item[] \textbf{Step 2}: We prove that the objective function can be approximated by a quadratic function around saddle point.
	\item[] \textbf{Step 3}: Combing \textbf{Step 1} and \textbf{Step 2}, we show the escape time bound for the objective function.
\end{itemize}

Now, let us start from the first step.  We construct the following quadratic function.
\begin{eqnarray}
\tilde{f}(\x)=f(\x_{k_0})+\nabla f(\x_{k_0})(\x-\x_{k_0})+\frac{1}{2}(\x-\x_{k_0})^T \nabla^2 f(\x_{k_0})(\x-\x_{k_0}).\nonumber
\end{eqnarray}
For simplicity, let $f(\x_{k_0})=f_0, \nabla f(\x_{k_0})=\nabla f_0$ and $\nabla^2 f(\x_{k_0})=\mathbf{H}_0$.
Therefore, we have 
\begin{eqnarray}
\nabla \tilde{f}(\x)=\nabla f_0+\mathbf{H}_0(\x-\x_{k_0}),\label{E1}\\
\nabla^2 \tilde{f} (\x)=\mathbf{H}_0.\label{E2}
\end{eqnarray}

From ASGLD, we have $\D_{k_0-2}=\sum_{j=0}^{k_0-2}\mathrm{Diag}\left\lbrace G(\x_j)\star G(\x_j)\right\rbrace+(k_0-1)\delta\mathrm{\bm I}$.
Using the quadratic function (\ref{E2}), we have that
\begin{eqnarray}
\tilde{\x}_{k+1}=\tilde{\x}_k-\eta \tilde{\D}_{k-1}^{-\alpha/2}\left\lbrace \nabla f_0+\mathbf{ H}_0(\tilde{\x}_k-\x_{k_0})\right\rbrace +\eta \tilde{\D}_{k-1}^{-\alpha/2}\tilde{\bm h}_k+\sqrt{2\eta}\tilde\D_{k-1}^{-\beta/2}\e_k, \label{dx}
\end{eqnarray}
where $\tilde{\bm h}_k=G(\x_k)-\nabla f(\x_k)$. From \textbf{Assumption 3}, we have $\mathbb{E}\left( \tilde{\bm h}_k\right)=0 $ and $\mathrm{Var}\left( \tilde{\bm h}_k\right)\leq  C/B_k $.
Equation (\ref{dx}) is equivalent to
\begin{eqnarray}
\tilde{\x}_{k+1}-\x_{k_0}=\left( \mathrm{\bm I}-\eta  \tilde{\D}_{k-1}^{-\alpha/2} \mathrm{\bm H}_0\right) \left( \tilde{\x}_k-\x_{k_0}\right)  +\eta \tilde{\D}_{k-1}^{-\alpha/2}\left( \tilde{\bm \zeta}_k+\nabla f_0\right),\label{diff2}
\end{eqnarray}
where $\tilde{\bm \zeta}_k=\tilde{\bm h}_k+\sqrt{2/\eta}\tilde\D_{k-1}^{(\alpha-\beta)/2}\e_k$. If $\alpha\leq \beta$, it is clearly true that
\begin{eqnarray}
\mathbb{E}\left( \tilde{\bm \zeta}_k\right)=0,\nonumber\\
\mathrm{Var}\left( \tilde{\bm \zeta}_k\right)=O(C/B_k),\nonumber
\end{eqnarray}
since $\tilde \D_{k-1}^{(\alpha-\beta)/2}=O(1)$ for $\alpha=\beta$ and $\tilde \D_{k-1}^{(\alpha-\beta)/2}\xrightarrow{}0$ for $\alpha<\beta$.
If $\beta<\alpha$,  $\eta \tilde{\D}_{k-1}^{-\alpha/2}\left( \tilde{\bm \zeta}_k+\nabla f_0\right)$ in (\ref{diff2}) can be rewritten as 
\begin{eqnarray}
\eta \tilde{\D}_{k-1}^{-\alpha/2}\left( \tilde{\bm \zeta}_k+\nabla f_0\right)=\eta \tilde{\D}_{k-1}^{-\beta/2}\left( \tilde{\D}_{k-1}^{-\alpha/2+\beta/2}+\sqrt{2/\eta}\e_k\right).\nonumber 
\end{eqnarray}
All rest proof procedures are same to $\alpha\leq \beta$.

Now, summing (\ref{diff2}) from $k_0+1$ to $k_0+T$,
we can obtain 
\begin{eqnarray}
&\tilde{\x}_{k_0+T}-\x_{k_0}&=\sum_{j=0}^{T-2}\eta \tilde{\D}_{k_0-1+j}^{-\alpha/2}\prod_{l=j}^{T-2} \left( \mathrm{\bm I}-\eta  \tilde{\D}_{k_0+l}^{-\alpha/2} \mathbf{H}_0\right)\left\lbrace \tilde{\bm \zeta}_{k_0+j}+\nabla f_0\right\rbrace \nonumber\\
&&+\eta\tilde{\D}_{k_0+T-2}^{-\alpha/2}\left\lbrace \tilde{\bm \zeta}_{k_0+k-1}+\nabla f_0\right\rbrace.\label{E3}
\end{eqnarray}

Taking the conditional expectation on (\ref{E3}), we have 
\begin{eqnarray}
&\mathbb{E}\left\lbrace \tilde{\x}_{k_0+T}-\x_{k_0}\mid \x_{k_0}\right\rbrace &=\sum_{j=0}^{T-2}\eta \mathbb{E}\left( \tilde{\D}_{k_0-1+j}^{-\alpha/2}\right) \prod_{l=j}^{T-2} \left( \mathrm{\bm I}-\eta \mathbb{E}\left(  \tilde{\D}_{k_0+l}^{-\alpha/2}\right)  \mathrm{\bm H}_0\right)\nabla f_0 \nonumber\\
&&+\eta\mathbb{E}\left( \tilde{\D}_{k_0+T-2}^{-\alpha/2}\right) \nabla f_0\nonumber\\
&&=\mathbf{M}(T)\nabla f_0.\label{E4}
\end{eqnarray}

Note that
\begin{eqnarray}
(k_0+j)\delta  \leq [\tilde{\D}_{k_0+j-1}]_{\ell\ell}\leq (k_0+j)\left( \delta +C+C_1\right) ,\nonumber
\end{eqnarray}
Thus, $ \tilde{\D}_{k_0+j-1}=O(k_0+j)$.
Let $\mathbf{H}_0=\mathbf{U}\Lambda \mathbf{ U}^T$.
Then, we can obtain that 
\begin{eqnarray}
\prod_{l=j}^{T-2} \left( \mathrm{\bm I}-\eta \mathbb{E}\left(  \tilde{\D}_{k_0+l}^{-\alpha/2}\right)  \mathbf{H}_0\right)=\mathbf U  \prod_{l=j}^{T-2} \left( \mathrm{\bm I}-\eta \mathbb{E}\left(  \tilde{\D}_{k_0+l}^{-\alpha/2}\right)  \Lambda\right) \mathbf U^T.\nonumber
\end{eqnarray}
Note that, we can rewrite  $\mathbf{ M}(T)=\mathbf{U} \mathbf{ O} \mathbf{ U}^T$ for some diagonal matrix $\mathbf O$.

Without loss of generality, we assume $\lambda_1\leq \lambda_2\leq \cdots \leq \lambda_d$, $\lambda_1=-\gamma<0$ and $\lambda_2\geq 0$. That is the worst case, where there exists only one negative eigenvalue.
We can show the following Lemma. 
\begin{lemma}
	\begin{eqnarray}
	\left(1+a_6\gamma \eta k^{-\alpha/2} \right)^{k-1-j} \leq \prod_{l=j}^{T-2} \left[  1+\gamma\eta O\left\lbrace (k_0+l)^{-\alpha/2}\right\rbrace  \right] \leq \left(  1+a_5\gamma \eta\right)^{k-1-j},\label{E5}
	\end{eqnarray}
	where $a_5, a_6$ are positive constants.
\end{lemma}

\begin{proof}
	SInce there exists a positive constant $a_5$, such that $\gamma\eta O\left\lbrace (k_0+l)^{-\alpha/2}\right\rbrace\leq a_5 \gamma\eta $., we have
	\begin{eqnarray}
	\prod_{l=j}^{T-2} \left[  1+\gamma\eta O\left\lbrace (k_0+l)^{-\alpha/2}\right\rbrace  \right] \leq \prod_{l=j}^{T-2}\left(  1+\gamma \eta a_5\right) = (1+a_5\gamma\eta)^{k-1-j}.\nonumber
	\end{eqnarray}
	For other side, $k_0+l\leq k_0+T$, which  leads to $(k_0+l)^{-\alpha/2}\geq (k_0+T)^{-\alpha/2} $. Then, 
	\begin{eqnarray}
	(k_0+T)^{-\alpha/2}=(k_0+T)^{-\alpha/2} k^{-\alpha/2} k^{\alpha/2}= \left(\frac{k_0+T}{T} \right)^{-\alpha/2} k^{-\alpha/2}\leq  k^{-\alpha/2}.\nonumber
	\end{eqnarray}
	Therefore, 
	\begin{eqnarray}
	\prod_{l=j}^{T-2} \left[  1+\gamma\eta O\left\lbrace (k_0+l)^{-\alpha/2}\right\rbrace  \right] \geq \prod_{l=j}^{T-2}\left( 1+a_6\gamma \eta k^{-\alpha/2}\right)=\left( 1+a_6\gamma \eta k^{-\alpha/2}\right)^{k-1-j}.\nonumber
	\end{eqnarray}
	Then, we complete the proof of (\ref{E5}).
\end{proof}

Similarly, for positive eigenvalues $\lambda$, we can establish that
\begin{eqnarray}
\left( 1-a_8\lambda \eta\right)^{k-1-j} \leq \prod_{l=j}^{T-2} \left[  1-\lambda\eta O\left\lbrace (k_0+l)^{-\alpha/2}\right\rbrace  \right]\leq \left( 1-a_7\lambda\eta k^{-\alpha/2}\right)^{k-1-j}<1. \label{E52}
\end{eqnarray}

The following claim can be established.
\begin{eqnarray}
O\left\lbrace (k_0-1)^{-\alpha/2}\right\rbrace \geq \tilde{\D}_{k_0-1+j}^{-\alpha/2}\geq \tilde{\D}_{k_0-1+T}^{-\alpha/2}=O\left\lbrace (k_0-1+T)^{-\alpha/2}\right\rbrace. \label{E53}
\end{eqnarray}

We then derive the second-order term. 
Using \textbf{Fact} \ref{fact}, we can show that
\begin{eqnarray}
\mathbb{E}\left\lbrace \left( \tilde{\x}_{k_0+T}-\x_{k_0}\right)^T\mathbf{ H}_0 \left( \tilde{\x}_{k_0+T}-\x_{k_0}\right)\mid \x_{k_0}\right\rbrace =\mathrm{Trace}\left\lbrace \mathbf H_0  \mathbf S \right\rbrace,\nonumber
\end{eqnarray}
where
\begin{eqnarray}
\mathbf S=\mathbb{E}\left\lbrace \left( \tilde{\x}_{k_0+T}-\x_{k_0}\right)\left( \tilde{\x}_{k_0+T}-\x_{k_0}\right)^T \mid \x_{k_0} \right\rbrace.
\end{eqnarray}
Note that, we can decompose $\mathbf S$ as
\begin{eqnarray}
&&\stackrel{\mathbf S=\underbrace{\mathbb{E}\left\lbrace \left( \tilde{\x}_{k_0+T}-\x_{k_0}\right) \mid \x_{k_0} \right\rbrace\mathbb{E}^T\left\lbrace \left( \tilde{\x}_{k_0+T}-\x_{k_0}\right) \mid \x_{k_0} \right\rbrace}}{J_1}\nonumber\\
&&\stackrel{+\underbrace{\mathrm{Var}\left\lbrace \left( \tilde{\x}_{k_0+T}-\x_{k_0}\right) \mid \x_{k_0} \right\rbrace}}{J_2}\nonumber.
\end{eqnarray}

For $J_1$, using (\ref{E4}), we have 
\begin{eqnarray}
J_1=\mathbf U \mathbf O \mathbf U^T\nabla f_0\left( \mathbf  U \mathbf O \mathbf U^T\nabla f_0\right)^T.
\end{eqnarray}
$\mathrm{Trace}(\mathbf{ H}_0 J_1)$ is 
\begin{eqnarray}
\mathrm{Trace}(\mathbf{H}_0 J_1)=\mathrm{Trace}\left\lbrace \left( \mathbf  U \mathbf O \mathbf U^T\nabla f_0\right)^T\mathbf{H}_0\mathbf U \mathbf O \mathbf U^T\nabla f_0\right\rbrace =\nabla f_0^T \mathbf U \mathbf O \Lambda \mathbf O \mathbf U^T \nabla f_0.\nonumber
\end{eqnarray}

For $J_2$ term, from (\ref{E3}), 
\begin{eqnarray}
&\tilde{\x}_{k_0+T}-\x_{k_0}&=\sum_{j=0}^{T-2}\eta \mathbf U\tilde{\D}_{k_0-1+j}^{-\alpha/2}\prod_{l=j}^{T-2} \left( \mathbf{ I}-\eta  \tilde{\D}_{k_0+l}^{-\alpha/2} \Lambda\right)\mathbf U^T\left\lbrace \tilde{\bm \zeta}_{k_0+j}+\nabla f_0\right\rbrace \nonumber\\
&&+\eta\tilde{\D}_{k_0+T-2}^{-\alpha/2}\left\lbrace \tilde{\bm \zeta}_{k_0+k-1}+\nabla f_0\right\rbrace.\nonumber
\end{eqnarray}
Then, we can show 
\begin{eqnarray}
J_2=O\left[ T\sum_{j=0}^{T-2}\eta^2 \tilde{\D}_{k_0-1+j}^{-\alpha}\prod_{l=j}^{T-2} \left( \mathrm{\bm I}-\eta  \tilde{\D}_{k_0+l}^{-\alpha/2} \Lambda\right)^2 +T\tilde{\D}_{k_0+T-2}^{-\alpha}\right] =d_1T\mathbf O\mathbf O,\nonumber
\end{eqnarray}
for some positive constant $d_1$.
Thus, 
\begin{eqnarray}
\mathrm{Trace}(\mathbf{ H}_0J_2)=d_2T\left\lbrace \mathrm{Trace}(\Lambda \mathbf O \mathbf O)\right\rbrace,\nonumber 
\end{eqnarray}
for some positive constant $d_2$.

To escape from the saddle point, we need to show that 
\begin{eqnarray}
\mathbb{E}\left\lbrace \tilde{f}(\x_{k_0+T})-f(\x_{k_0})\mid \x_{k_0}\right\rbrace< 0.\nonumber 
\end{eqnarray}
Note that,
\begin{eqnarray}
&\mathbb{E}\left\lbrace \tilde{f}(\x_{k_0+T})-f(\x_{k_0})\mid \x_{k_0}\right\rbrace&=\nabla f_0^T \mathbb{E}\left\lbrace \x_{k_0+T}-\x_{k_0}\mid \x_{k_0}\right\rbrace\nonumber\\
&& +\frac{1}{2} \mathbb{E}\left\lbrace \left( \x_{k_0+T}-\x_{k_0} \right)^T \mathbf{ H}_0 \left(  \x_{k_0+T}-\x_{k_0}\mid \x_{k_0}\right) \right\rbrace.\nonumber
\end{eqnarray}
That is equivalent to showing
\begin{eqnarray}
\nabla^T f_0 \mathbf{ M}(T)\nabla f_0 +\frac{1}{2}\mathrm{Trace}(\mathbf{ H}_0\mathbf S) < 0.\label{E9}
\end{eqnarray}

Thus, to show (\ref{E9}) is equivalent to showing
\begin{eqnarray}
\nabla^T f_0 \mathbf U \mathbf O \mathbf U^T\nabla f_0+\nabla f_0^T \mathbf U \mathbf O \Lambda \mathbf O \mathbf U^T \nabla f_0+d_2 T\mathrm{Trace}(\Lambda \mathbf O \mathbf O)<0.\label{term2}
\end{eqnarray}
Let $\mathbf O=\mathrm{Diag}\left\lbrace (O_1, \cdots O_d)\right\rbrace $ and $\mathbf U^T\nabla f_0=(f_1, \cdots , f_d)$.
Then, if $\|\nabla f_0\|\leq \epsilon$. We have $\|\mathbf U^T\nabla f_0\|\leq \epsilon$.
We can rewrite (\ref{term2}) as
\begin{eqnarray}
\sum_{j=1}^d O_j f_j^2 +\sum_{j=1}^d \lambda_j O_j ^2f_j^2 +d_2T\sum_{j=1}^d \lambda_j O_j^2.\label{term3} 
\end{eqnarray}

Using (\ref{E52}) and \ref{E53}, we can conclude that $0<O_j<1$, for $j=2,3,\cdots, d$. Similarly, we can show that 
\begin{eqnarray}
(1+a_6\gamma \eta k^{-\alpha/2})^{k-1}(k_0-1+T)^{-\alpha/2}\leq O_1\leq (1+a_5\gamma \eta)^{k-1} (k_0-1)^{-\alpha/2}.\nonumber
\end{eqnarray}
Therefore, 
\begin{eqnarray}
\sum_{j=1}^d \lambda_j O_j^2\leq -\gamma O_1^2+ d\sum_{j=2}\lambda_j\label{cond1}.
\end{eqnarray}
Moreover, note that
\begin{eqnarray}
\sum_{j=1}^d O_j f_j^2 +\sum_{j=1}^d \lambda_j O_j ^2f_j^2\leq f_1^2 (O_1-\gamma O_1^2)+ d\sum_{j=2}^d\lambda_j (\epsilon^2 -f_1^2)\leq -\frac{\gamma}{2}f_1^2 O_1^2+ d\sum_{j=2}^d\lambda_j (\epsilon^2 -f_1^2),\label{cond2}
\end{eqnarray}
for large $T$.
Combing (\ref{cond1}) and (\ref{cond2}) yields:
\begin{eqnarray}
-\gamma O_1^2 (d_2T+f_1^2)+d\sum_{j=2}^2\lambda_j(d_2T+\epsilon^2 -f_1^2)<0.\nonumber
\end{eqnarray}
Since $\epsilon^2$ is negligible comparing to $d_2T$, it is sufficient to show 
\begin{eqnarray}
-\gamma O_1^2 +d\sum_{j=2}^2\lambda_j<0.\label{cond3}
\end{eqnarray}
Since $O_1$ increases exponentially w.r.t. $T$, (\ref{cond3}) holds for sufficiently large $T$. 
That is, as long as $T$ is large enough, we have 
\begin{eqnarray}
\mathbb{E}\left\lbrace \tilde{f}(\x_{k_0+T})-f(\x_{k_0})\mid \x_{k_0}\right\rbrace< 0.\nonumber 
\end{eqnarray}

To compute $T_{\max}$, it is sufficient to let 
\begin{eqnarray}
-\gamma O_1^2 +d\sum_{j=2}^2\lambda_j\leq -\gamma (1+a_6\gamma \eta k^{-\alpha/2})^{2T-2}(k_0-1+T)^{-\alpha} +d\sum_{j=2}^2\lambda_j<0.
\end{eqnarray}
Then, we can obtain that $T=O\left( (a_7\log d+a_9 \alpha) /(\log(\gamma)+a8)\right) $, where $a_7, a_8$ are positive numbers. Similarly, for $\beta<\alpha$, we have $T=O\left( (a_7\log d+a_9 \beta) /(\log(\gamma)+a8)\right) $. For finite $\alpha, \beta$, we can conclude that at most $T_{\max}=O(\log d)$ iterations are used to escape from $\gamma$-strict saddle point.
Then, we can completes our proof for \textbf{Step 1}.


At \textbf{Step 2}, we will show that the objective function can be approximated by the quadratic function. Note that, 
\begin{eqnarray}
\tilde{\x}_{k_0+1}={\x}_{k_0}-\eta \tilde{\D}_{k_0-1}^{-\alpha/2}\left\lbrace \nabla f_0+\mathbf{H}_0({\x}_{k_0}-\x_{k_0})\right\rbrace +\eta \tilde{\D}_{k_0-1}^{-\alpha/2}{\bm \zeta}_{k_0}, \nonumber
\end{eqnarray}
and 
\begin{eqnarray}
{\x}_{k_0+1}=\tilde{\x}_{k_0}-\eta{ \D}_{k_0-1}^{-\alpha/2} \nabla f_0 +\eta{\D}_{k_0-1}^{-\alpha/2}{\bm \zeta}_{k_0}. \nonumber
\end{eqnarray}
Thus, we have $\x_{k_0+1}=\tilde{\x}_{k_0+1}$.
Furthermore, note that
\begin{eqnarray}
&\|\x_{k_0+j}-\tilde{\x}_{k_0+j}\|&\leq \eta O\left\lbrace (k_0+j)^{-\alpha/2}\right\rbrace \left\| \nabla f(\x_{k_0+j-1})-\nabla \tilde{f}(\tilde{\x}_{k_0+j-1})\right\|\nonumber\\
&& +\eta O\left\lbrace (k_0+j)^{-\alpha/2}\right\rbrace\|\bm \zeta_{k_0+j-1}-\tilde{\bm \zeta}_{k_0+j-1} \|.\nonumber 
\end{eqnarray}
Taking conditional expectation with respect to $\x_{k_0+j-1}$ and $\tilde{\x}_{k_0+j-1}$, we have 
\begin{eqnarray}
&\left\|\x_{k_0+j-1}-\tilde{\x}_{k_0+j-1} \right\|&\leq  \eta O\left\lbrace (k_0+j)^{-\alpha/2}\right\rbrace \left\| \nabla f(\x_{k_0+j-1})-\nabla \tilde{f}(\tilde{\x}_{k_0+j-1})\right\|\nonumber\\
&&+\eta O\left\lbrace (k_0+j)^{-\alpha/2}\right\rbrace,\nonumber
\end{eqnarray}
since 
\begin{eqnarray}
\mathbb{E}\left\lbrace \|\bm \zeta_{k_0+j-1}-\tilde{\bm \zeta}_{k_0+j-1} \|\right\rbrace \leq \left[ \mathbb{E}\left\lbrace \|\bm \zeta_{k_0+j-1}-\tilde{\bm \zeta}_{k_0+j-1} \|^2\right\rbrace\right]^{1/2} \leq C^{1/2}.\nonumber
\end{eqnarray}

Moreover, since
\begin{eqnarray}
\left\|\nabla f(\x_{k_0+j-1})-\nabla \tilde{f}(\tilde{\x}_{k_0+j-1}) \right\| \leq \frac{\rho}{2}\left\| \x_{k_0+j-1}-\tilde{\x}_{k_0+j-1}\right\|^2,\nonumber
\end{eqnarray}
we have:
\begin{eqnarray}
&\|\x_{k_0+j}-\tilde{\x}_{k_0+j}\|&\leq \eta O\left\lbrace (k_0+j)^{-\alpha/2}\right\rbrace \frac{\rho}{2}\left\| \x_{k_0+j-1}-\tilde{\x}_{k_0+j-1}\right\|^2\nonumber\\
&& +\eta O\left\lbrace (k_0+j)^{-\alpha/2}\right\rbrace\nonumber\\
&&\leq \sum_{l=0}^{j-1}\eta^{j-l} O\left\lbrace (k_0-1+l)^{-(j-l)\alpha/2}\right\rbrace\nonumber\\
&&\leq \frac{1-\eta^jO\left\lbrace (k_0-1)^{-j\alpha/2}\right\rbrace}{1-\eta} \leq \frac{1}{1-\eta}.\label{diff}
\end{eqnarray}

Finally, for the objective function, we have 
\begin{eqnarray}
&f(\x_{k_0+T})-f(\x_{k_0})&\leq \nabla f_0(\x_{k_0+T}-\x_{k_0})+\frac{1}{2}(\x_{k_0+T}-\x_{k_0})^T\mathrm{\bm H}_0(\x_{k_0+T}-\x_{k_0})\nonumber\\
&&+\frac{\rho}{6}\|\x_{k_0+T}-\x_{k_0}\|^3.\nonumber
\end{eqnarray}
Letting  $(\x_{k_0+T}-\x_{k_0})=(\x_{k_0+T}-\tilde{\x}_{k_0})+(\tilde{\x}_{k_0+T}-\x_{k_0})=\Delta_1+\Delta_2$,
we then have:
\begin{eqnarray}
&f(\x_{k_0+T})-f(\x_{k_0})&\leq \nabla f_0(\Delta_1+\Delta_2)+\frac{1}{2}(\Delta_1+\Delta_2)^T\mathbf{ H}_0 (\Delta_1+\Delta_2)\nonumber\\
&&+\frac{\rho}{6}\|\Delta_1+\Delta_2\|^3\nonumber\\
&&=\nabla f_0\Delta_2+\frac{1}{2}\Delta_2^T\mathbf{ H}_0\Delta_2 \nonumber\\
&&+\nabla f_0\Delta_1+\frac{1}{2}\Delta_1^T\mathbf{ H}_0 \Delta_1+\Delta_1^T\mathbf{ H}_0\Delta_2+\frac{\rho}{6}\|\Delta_1+\Delta_2\|^3.\label{final}
\end{eqnarray}
We already show that 
\begin{eqnarray}
\mathbb{E}\left\lbrace \nabla f_0\Delta_2+\frac{1}{2}\Delta_2^T\mathbf{ H}_0\Delta_2\mid \x_{k_0}\right\rbrace<0,\nonumber 
\end{eqnarray}
when $T$ is large enough. 
From (\ref{diff}), we have $\|\Delta_1\|\leq 1$.
Then, $\|\nabla f_0 \Delta_1\|\leq \|\Delta_1\|\|\nabla f_0\|\leq \epsilon$. Moreover, 
\begin{eqnarray}
\frac{1}{2}\Delta_1^T \mathbf{H}_0\Delta_1\leq \frac{1}{2}\Delta_1^T\Delta_1 \mathrm{Trace}(\mathbf{ H}_0)=O(d).\label{cond4}
\end{eqnarray}

Note that, we have 
\begin{eqnarray}
\mathbb{E}\left\lbrace \|\tilde{\x}_{k_0+1}-\tilde{\x}_{k_0}\|^2\right\rbrace\leq \frac{\eta^2M}{2}\nabla f_0^T \mathbb{E}\left(\tilde{\D}_{k_0-1}^{-\alpha} \right)  \nabla f_0+\frac{\eta^2M}{2}\mathrm{Trace}\left\lbrace  \mathbb{E}\left(\tilde{\D}_{k_0-1}^{-\alpha} \right)  \right\rbrace C/B_k=O(k_0^{-\alpha}).\nonumber
\end{eqnarray}
Therefore, 
\begin{eqnarray}
\mathbb{E}\left\lbrace \|\tilde{\x}_{k_0+1}-\tilde{\x}_{k_0}\|^2\right\rbrace\leq \sum_{j=0}^{T}O\left\lbrace (k_0+j)^{-\alpha}\right\rbrace \leq O\left\lbrace (k_0+T)^{1-\alpha}\right\rbrace. \nonumber
\end{eqnarray}
That is 
\begin{eqnarray}
\mathbb{E}\|\Delta_2\|^2\leq O\left\lbrace (k_0+T)^{1-\alpha}\right\rbrace.\nonumber
\end{eqnarray}
Then, 
\begin{eqnarray}
\mathbb{E}\left( \Delta_1^T\mathbf{ H}_0\Delta_2\right) \leq \mathbb{E}\left( \|\Delta_1^T\mathbf{H}_0\Delta_2\|\right) \leq \mathbb{E}\left( \|\Delta_1\|\Delta_2\|\right) \mathrm{Trace}(\mathbf{H}_0)\leq O\left\lbrace d  (k_0+T)^{(1-\alpha)/2}\right\rbrace.\label{cond5}
\end{eqnarray}
For the last term,  
\begin{eqnarray}
\left( \mathbb{E}\left\lbrace \|\Delta_1+\Delta_2\|^3\right\rbrace\right)^{1/3} \leq\left( \mathbb{E}\left\lbrace \|\Delta_1\|^3\right\rbrace\right)^{1/3}+\left( \mathbb{E}\left\lbrace \|\Delta_2\|^3\right\rbrace\right)^{1/3}.\nonumber
\end{eqnarray}
Since $\|\Delta_1\|\leq 1$, $\left( \mathbb{E}\left\lbrace \|\Delta_1\|^3\right\rbrace\right)^{1/3}\leq 1$.
Note that, we can rewrite
\begin{eqnarray}
\|\bm A\|^3=\sum_{j=1}^d =\left( \sum_{l_1, l_2,l_3} A_{l_1}^2 A^2_{l_2}A^2_{l_3}\right)^{1/2},\nonumber
\end{eqnarray}
where $\bm A=(A_1, \cdots A_d)^T$. Then,
\begin{eqnarray}
\mathbb{E}\left\lbrace \|\tilde{\x}_{k_0+1}-\tilde{\x}_{k_0}\|^3\right\rbrace\leq O\left[  \mathrm{Trace}\left\lbrace \mathbb{E}\left(\tilde{\D}_{k_0-1}^{-3\alpha/2} \right)\right\rbrace d^3\right]=O\left( k_0^{-3\alpha/2} d^3 \right). \label{cond6}
\end{eqnarray}
Our target is to show 
\begin{eqnarray}
\mathbb{E}\left\lbrace f(\x_{k_0+T})-f(\x_{k_0})\mid \x_{k_0}\right\rbrace <0.\nonumber
\end{eqnarray}
Therefore, using (\ref{cond3}), (\ref{cond4}), (\ref{cond5}) and (\ref{cond6}), it is sufficient to show 
\begin{multline}
-\gamma (1+a_6\gamma \eta k^{-\alpha/2})^{2T-2}(k_0-1+T)^{-\alpha} +d\sum_{j=2}^2\lambda_j+\epsilon \\
+O(d)+O(d(k_0+T)^{1-\alpha})+O\left( k_0^{-3\alpha/2} d^3 \right)<0.\label{f2}
\end{multline}
We can see that the negative term in (\ref{f2}) goes to infinity exponentially fast as $T$ increases. Thus, as long as $T$ is sufficiently large, we have that (\ref{f2}) is negative, which  implies that 
\begin{eqnarray}
\mathbb{E}\left\lbrace f(\x_{k_0+T})-f(\x_{k_0})\mid \x_{k_0}\right\rbrace <0.\nonumber
\end{eqnarray}
Therefore, $O(\log d)$ is the approximate maximum escape time, since the polynomial function of $T$ is negligible comparing to the exponential function in (\ref{f2}).

\subsection{Proof for AGLD}
The proof for AGLD is almost the same as ASGLD, with difference appearing in
\begin{eqnarray}
\tilde{\bm \zeta}_k=\tilde{\bm h}_t+\sqrt{2/\eta}\D_{k-1}^{(\alpha-\beta)/2}\e_k\nonumber
\end{eqnarray}
Now, we define 
\begin{eqnarray}
\tilde{\bm \zeta}_k=\sqrt{2/\eta}\D_{k-1}^{(\alpha-\beta)/2}\e_k
\end{eqnarray}
The rest of the proof follows  exactly the same steps as in proof for ASGLD.

\section{Experiment}
In this section, we conduct simulation studies to verify the performance of the proposed algorithms.  Let $\{\bm y_i=(y_{i1}, y_{i2})\}_{i=1}^n$ be $n$ independent and identically distributed realizations from  a multivariate Gaussian distribution with mean $(0,0)$ and variance $\mathrm{Diag}(0.1, 10)$. Let the sample size be $n=10,000$.  The parameter of interest is the variance. Suppose that we know the mean is $(0,0)$ and the off-diagonal entries in variance are 0. Our objective function is 
\begin{eqnarray}
f(\x)=\sum_{i=1}^n\left(  \log x_1 +\log x_2 +\frac{y_{i1}^2}{x_1}+\frac{y_{i2}^2}{x_2}\right) \label{obj2}
\end{eqnarray}
where $\x=(x_1, x_2)$ is positive. Note that, the objective function in (\ref{obj2}) is nonconvex. The parameter space is bounded below from 0. Two parameters have different magnitudes. Thus, the injection of random noise has to be chosen properly in SGLD to persist parameter space.
To minimize (\ref{obj2}) respect to $\x$, we apply the following methods.
\begin{itemize}
	\item [1.] SGLD: For each iteration, randomly pick one sample to construct the stochastic gradient.
	\item[2.] SGLD\_B: Randomly select $B=10$ samples to construct the stochastic gradient for each iteration.
	\item[3.] SGLD\_A: Randomly select $B=10$ samples to construct the stochastic gradient for each iteration. The step size is adaptively $\eta_t=(1+t)^{-1}$.
	\item[4.] ASGLD: Let $\alpha=1, \beta=1$. For each iteration, randomly pick one sample to construct the stochastic gradient.
	\item[5.] ASGLD\_B: Use $\alpha=1, \beta=1$. For each iteration,  select $B=10$ samples to construct the stochastic gradient randomly.
	\item[6.] ASGLD\_I: Use $\alpha=0, \beta=1$. For each iteration,  select $B_k=k$ samples to construct the stochastic gradient randomly.
	\item[7.] ASGLD2: Use $\alpha=1, \beta=2$. For each iteration,  select $B=10$ samples to construct the stochastic gradient randomly.
	\item[8.] ASGLD3: Use $\alpha=1, \beta=0.3$. For each iteration,  select $B=10$ samples to construct the stochastic gradient randomly.
	\item[9.] ASG: Use $\alpha=1$. Randomly select one sample to construct the stochastic gradient.
	\item[10.] AGLD: Use $\alpha=0, \beta=1$.	
	\item[11.] AGLD: Use $\alpha=0, \beta=2$.
	\item[12.] AGLD: Use $\alpha=0, \beta=0.3$.
\end{itemize}
To compare these method, we define the following evaluation:
\begin{eqnarray}
\text{error}= (x_{i1}-0.1)^2+(x_{i2}-10)^2.\nonumber
\end{eqnarray}

\begin{figure}
	\centering
	\begin{subfigure}[b]{0.4\textwidth}
		\includegraphics[width=\textwidth]{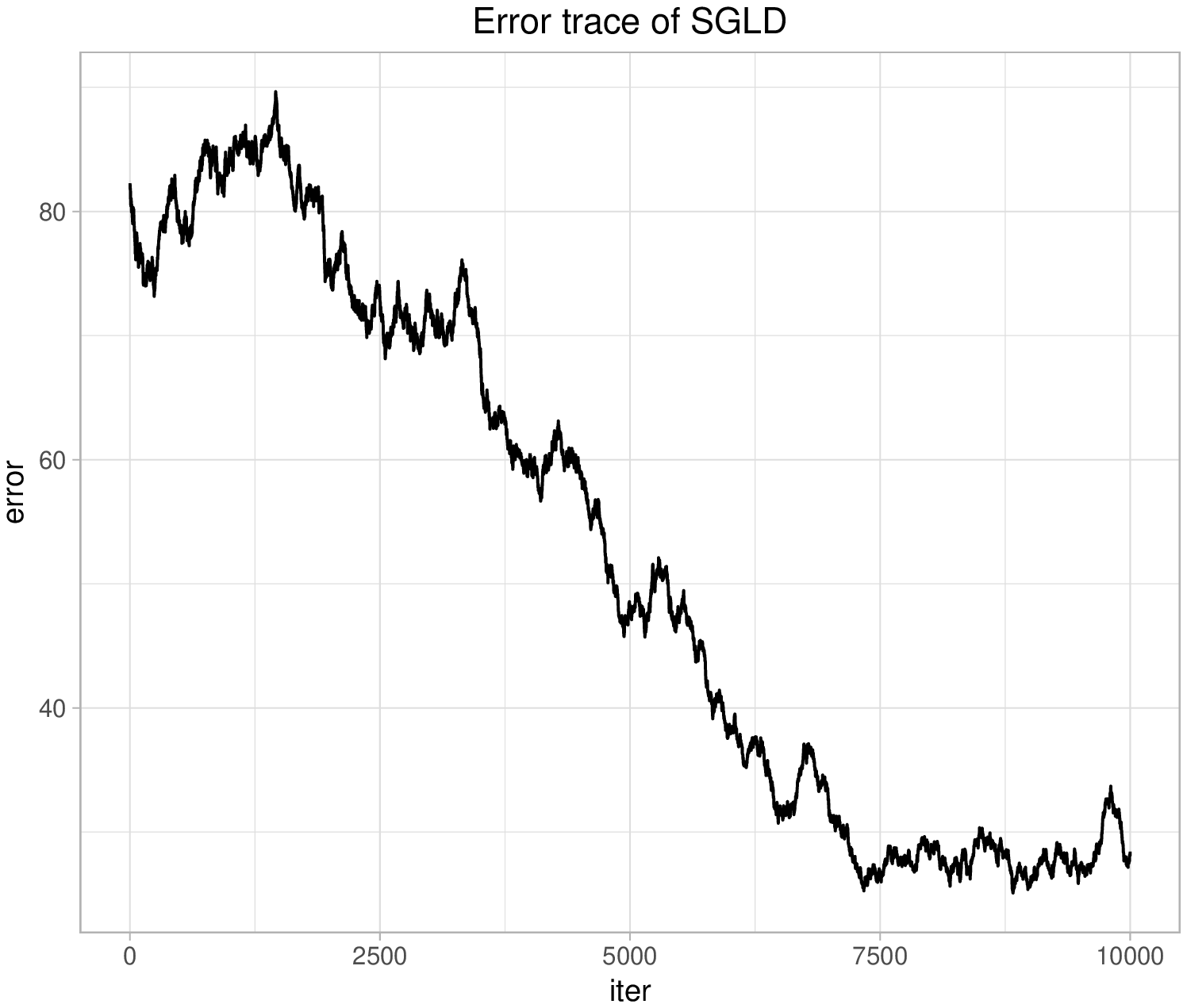}
		\caption{Error for SGLD}
	\end{subfigure}
	~ 
	\begin{subfigure}[b]{0.4\textwidth}
		\includegraphics[width=\textwidth]{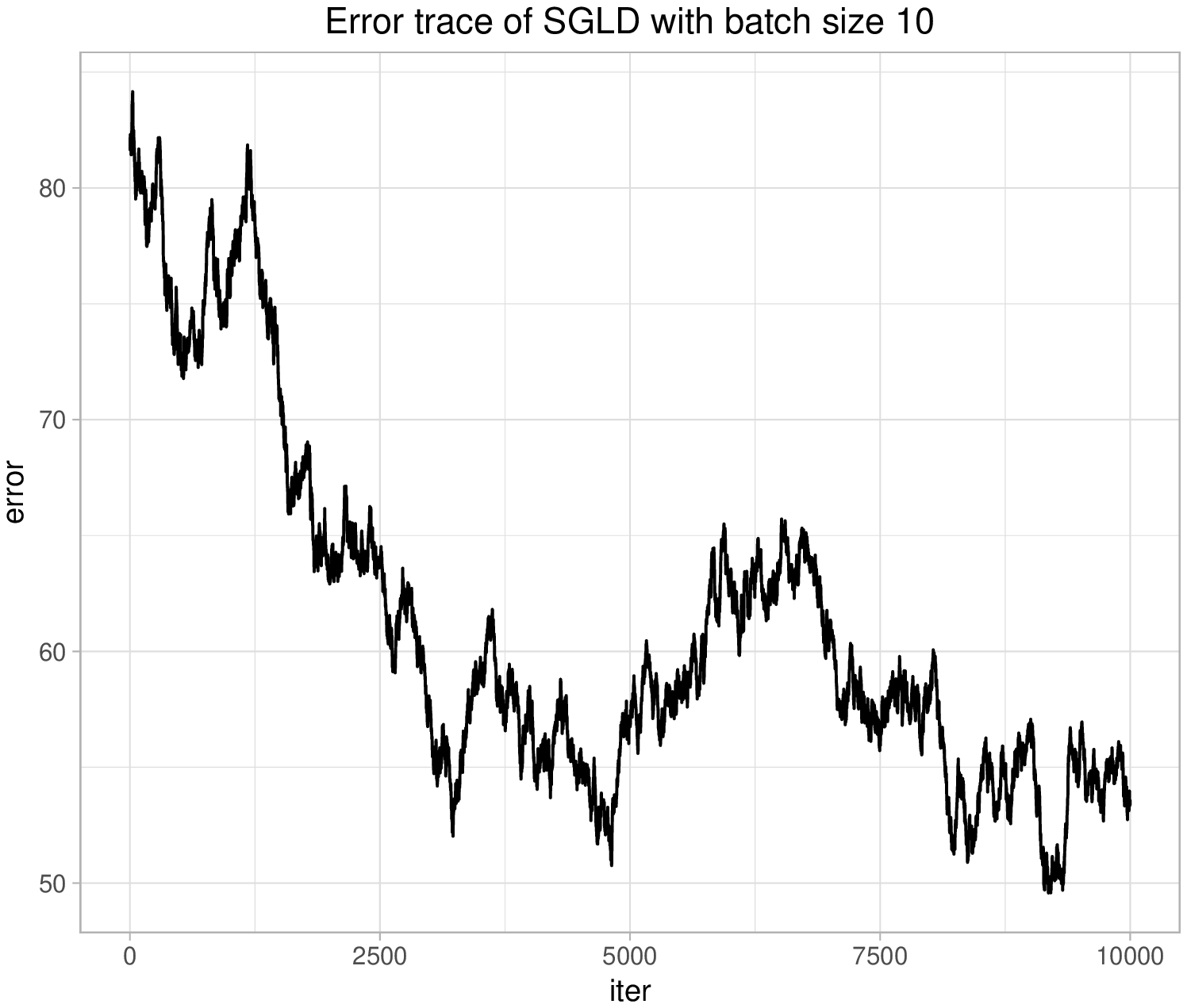}
		\caption{Error for SGLD using batch size 10}
	\end{subfigure}
	~ 
	\begin{subfigure}[b]{0.4\textwidth}
		\includegraphics[width=\textwidth]{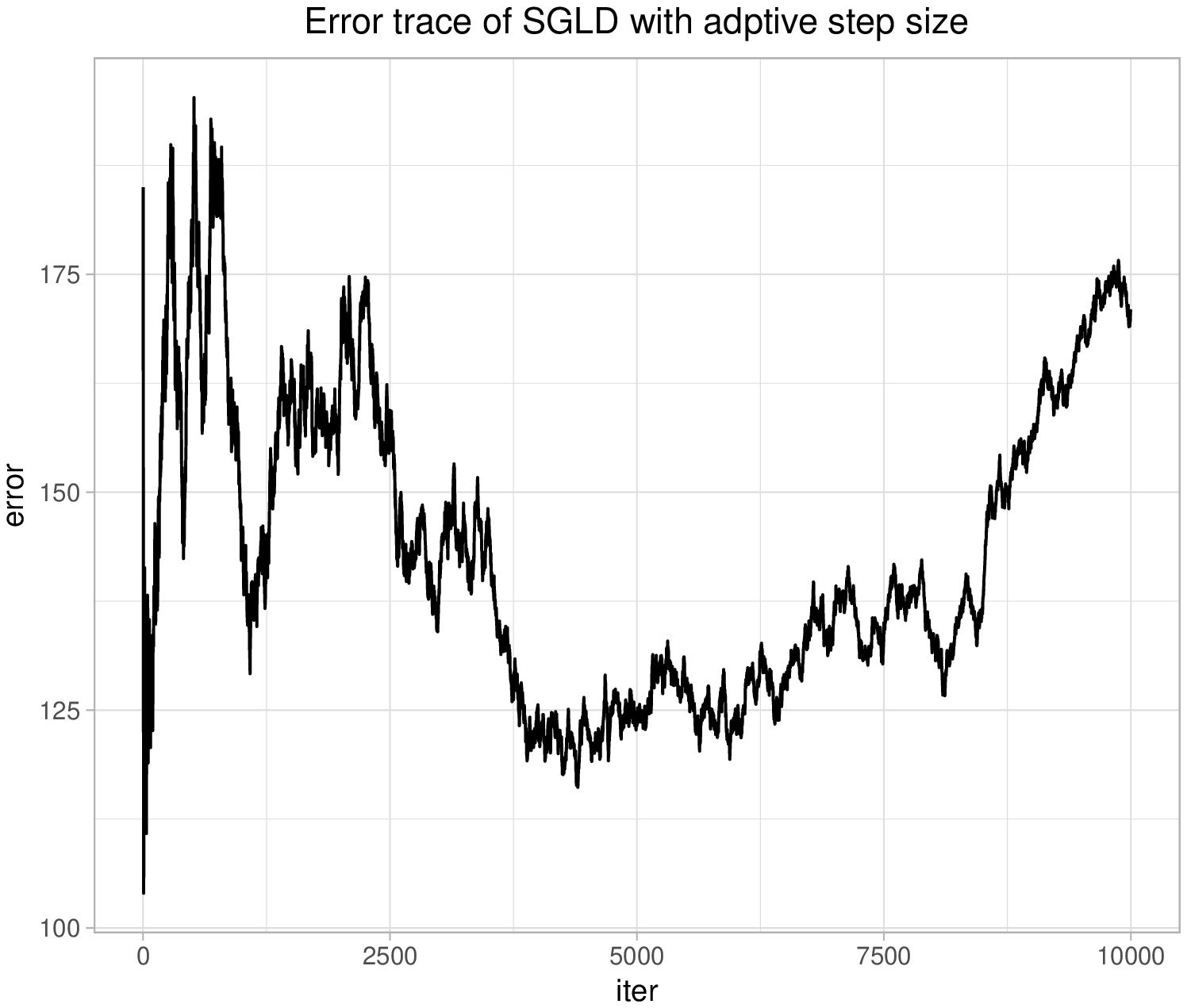}
		\caption{Error for SGLD using adaptive step size}
	\end{subfigure}
	~ 
	\begin{subfigure}[b]{0.4\textwidth}
		\includegraphics[width=\textwidth]{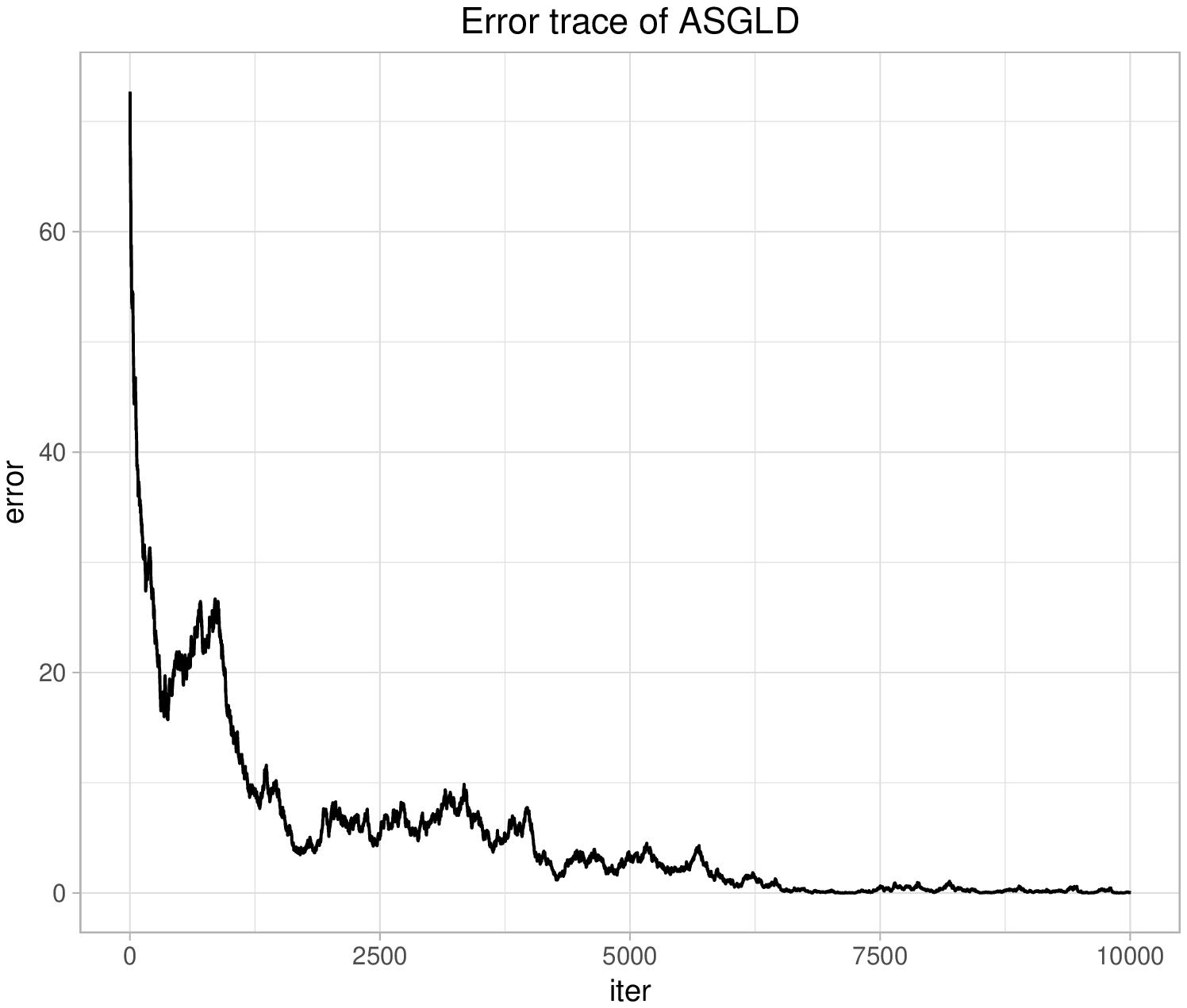}
		\caption{Error for ASGLD}
	\end{subfigure}
	\begin{subfigure}[b]{0.4\textwidth}
		\includegraphics[width=\textwidth]{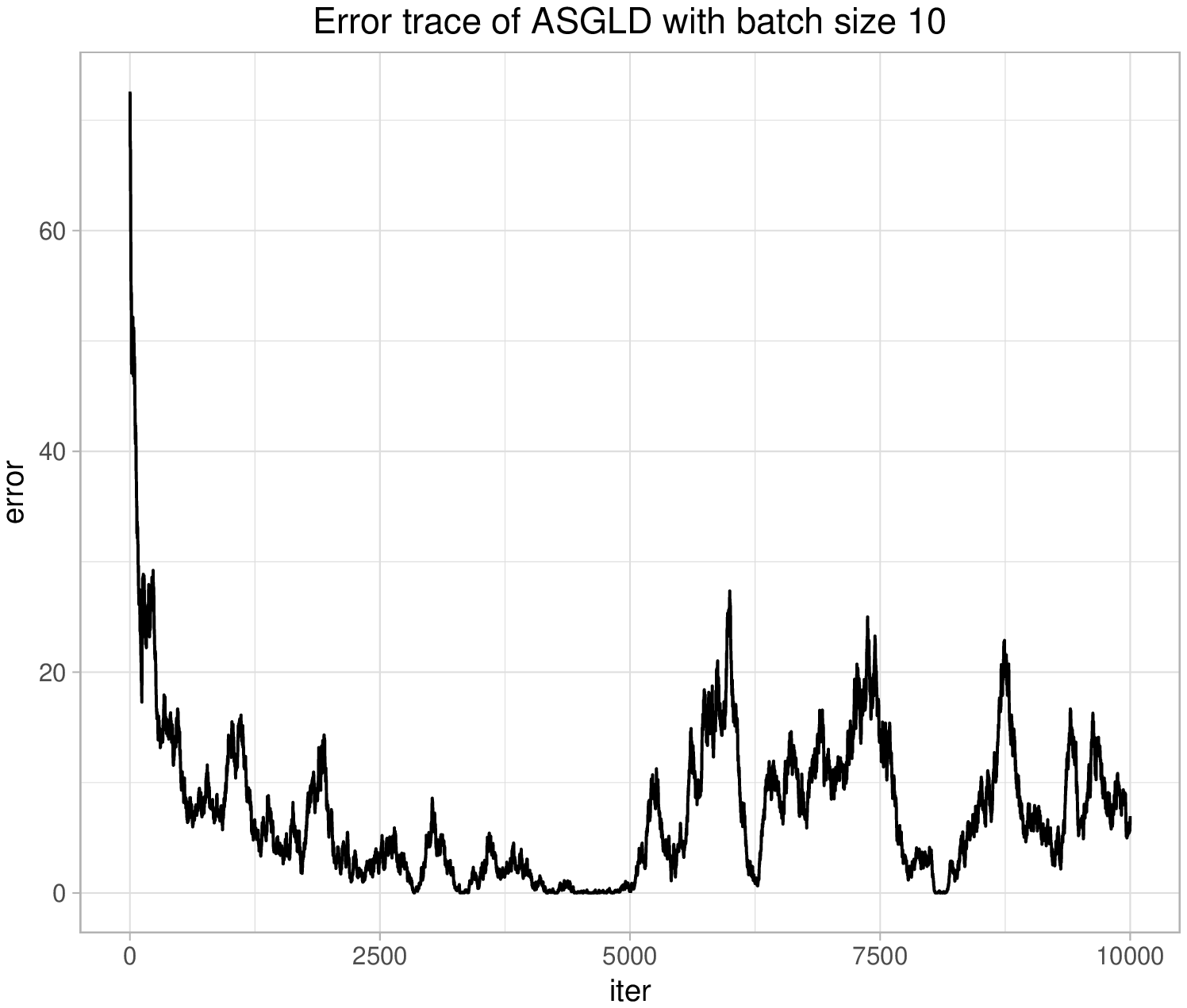}
		\caption{Error for ASGLD using batch step size 10.}
	\end{subfigure}
	~ 
	\begin{subfigure}[b]{0.4\textwidth}
		\includegraphics[width=\textwidth]{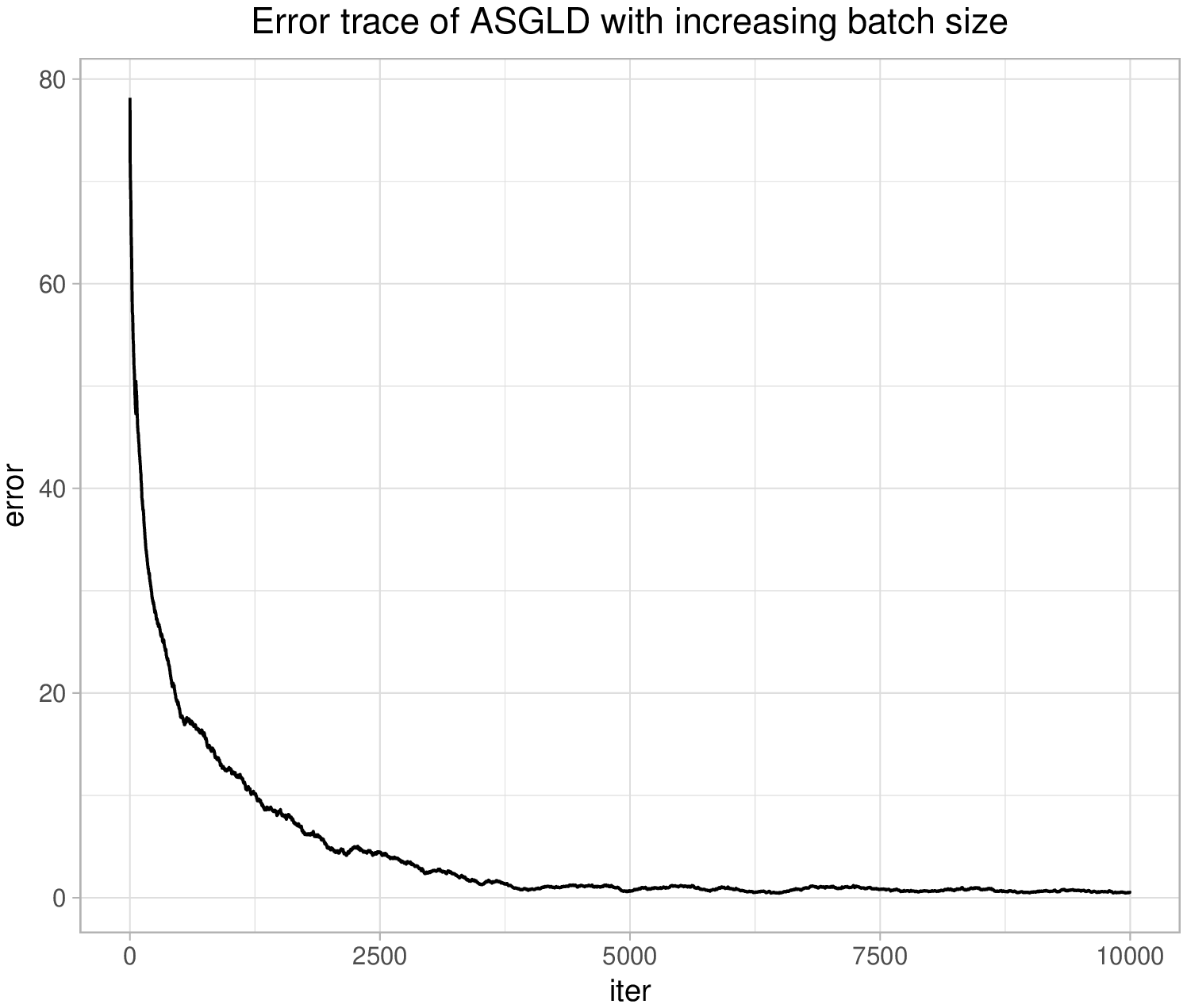}
		\caption{Error for ASGLD using increasing batch size.}
	\end{subfigure}
	\caption{Simulation results I.}\label{fig11}
\end{figure}

\begin{figure}
	\centering
	\begin{subfigure}[b]{0.4\textwidth}
		\includegraphics[width=\textwidth]{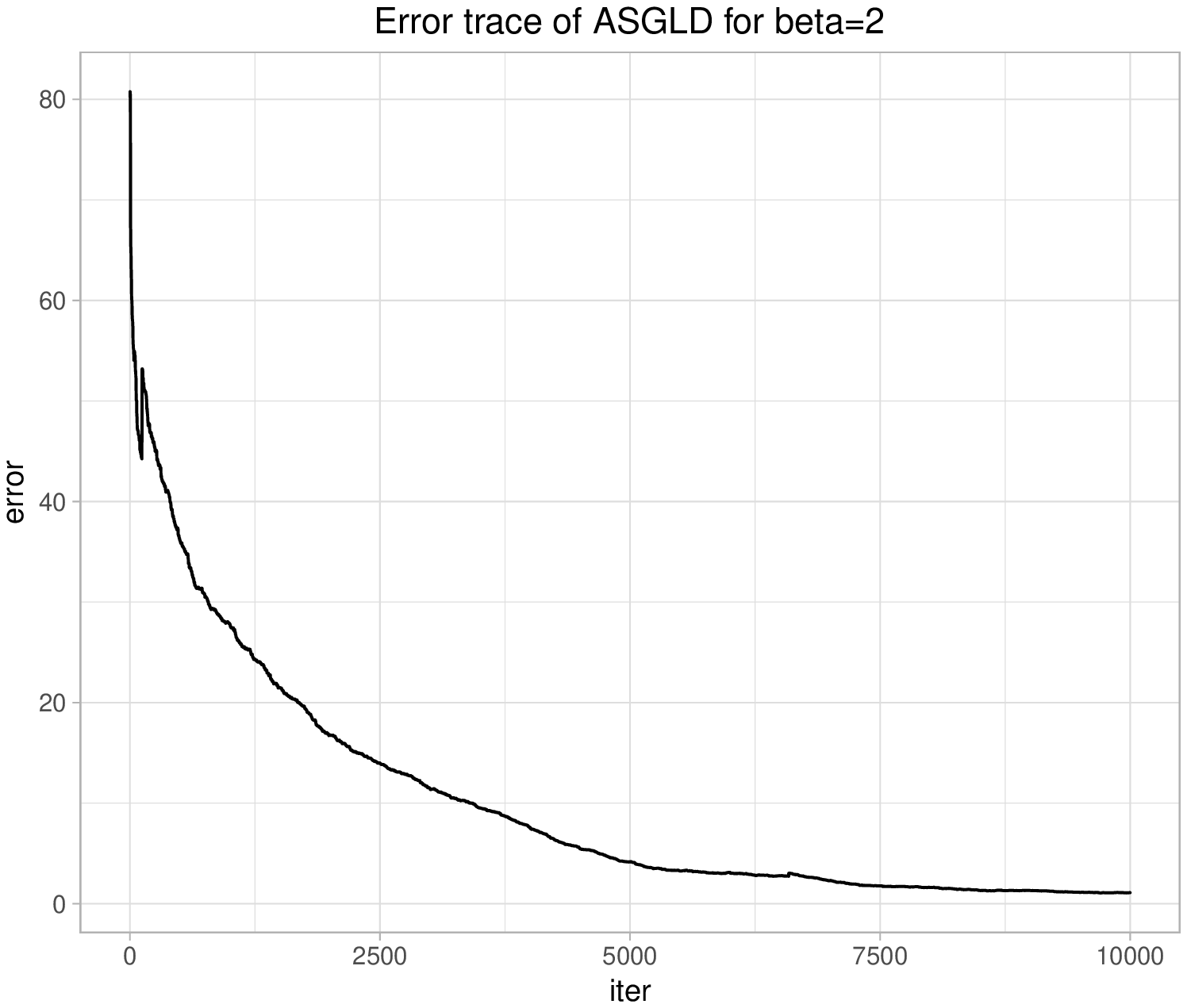}
		\caption{Error for ASGLD using $\beta=2$.}
	\end{subfigure}
	~ 
	\begin{subfigure}[b]{0.4\textwidth}
		\includegraphics[width=\textwidth]{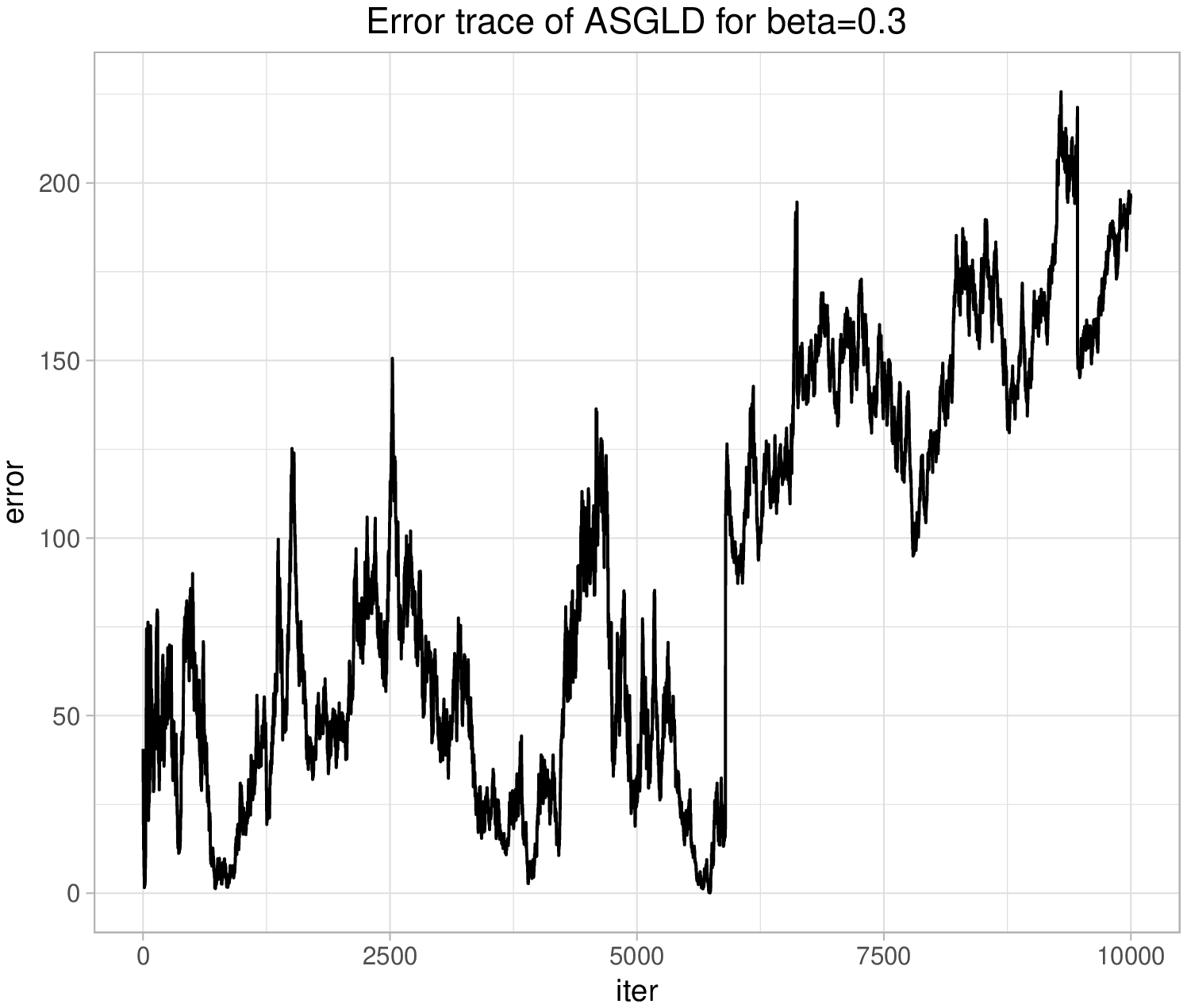}
		\caption{Error for ASGLD using $\beta=0.3$.}
	\end{subfigure}
	~ 
	\begin{subfigure}[b]{0.4\textwidth}
		\includegraphics[width=\textwidth]{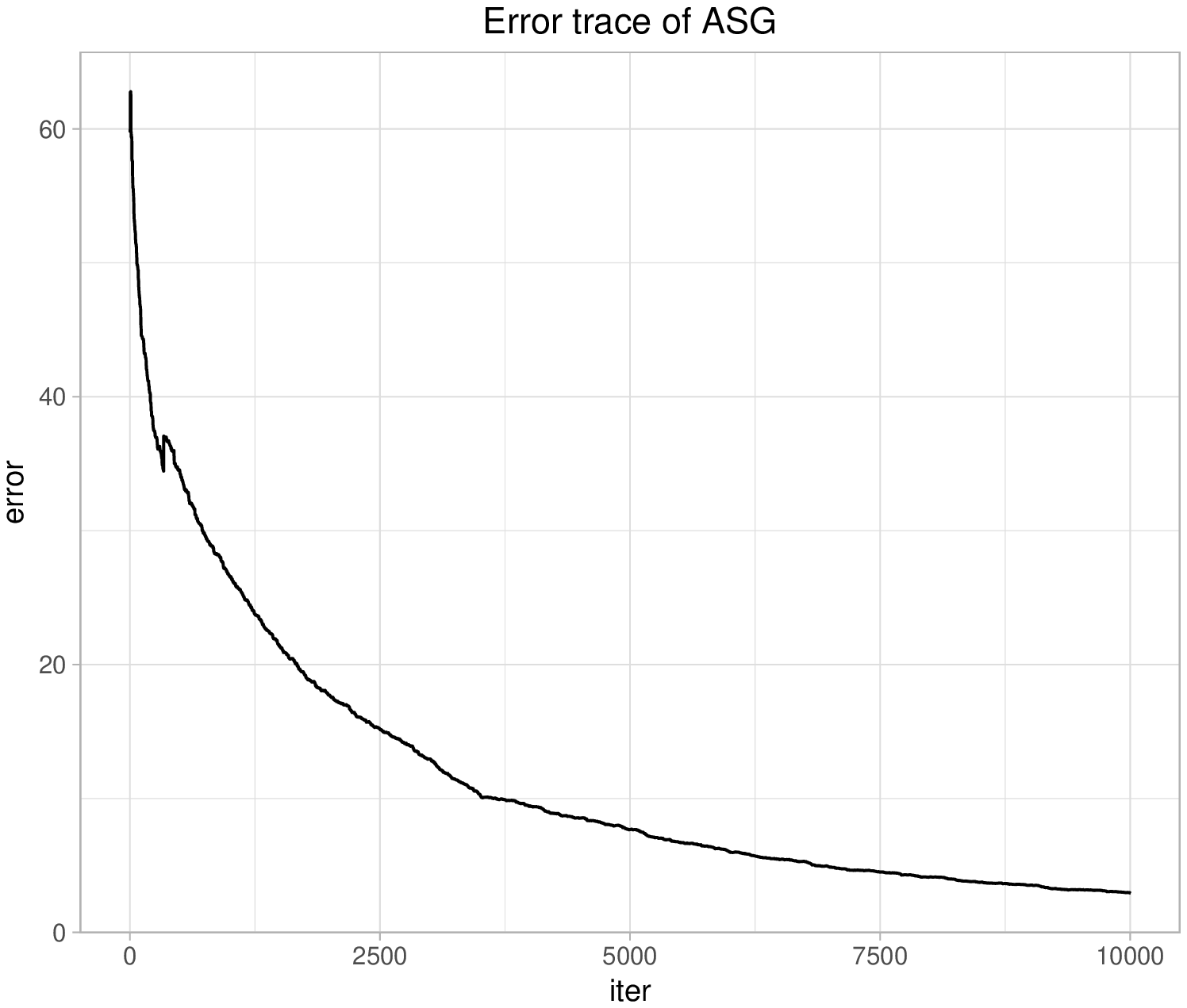}
		\caption{Error for ASG.}
	\end{subfigure}
	~ 
	\begin{subfigure}[b]{0.4\textwidth}
		\includegraphics[width=\textwidth]{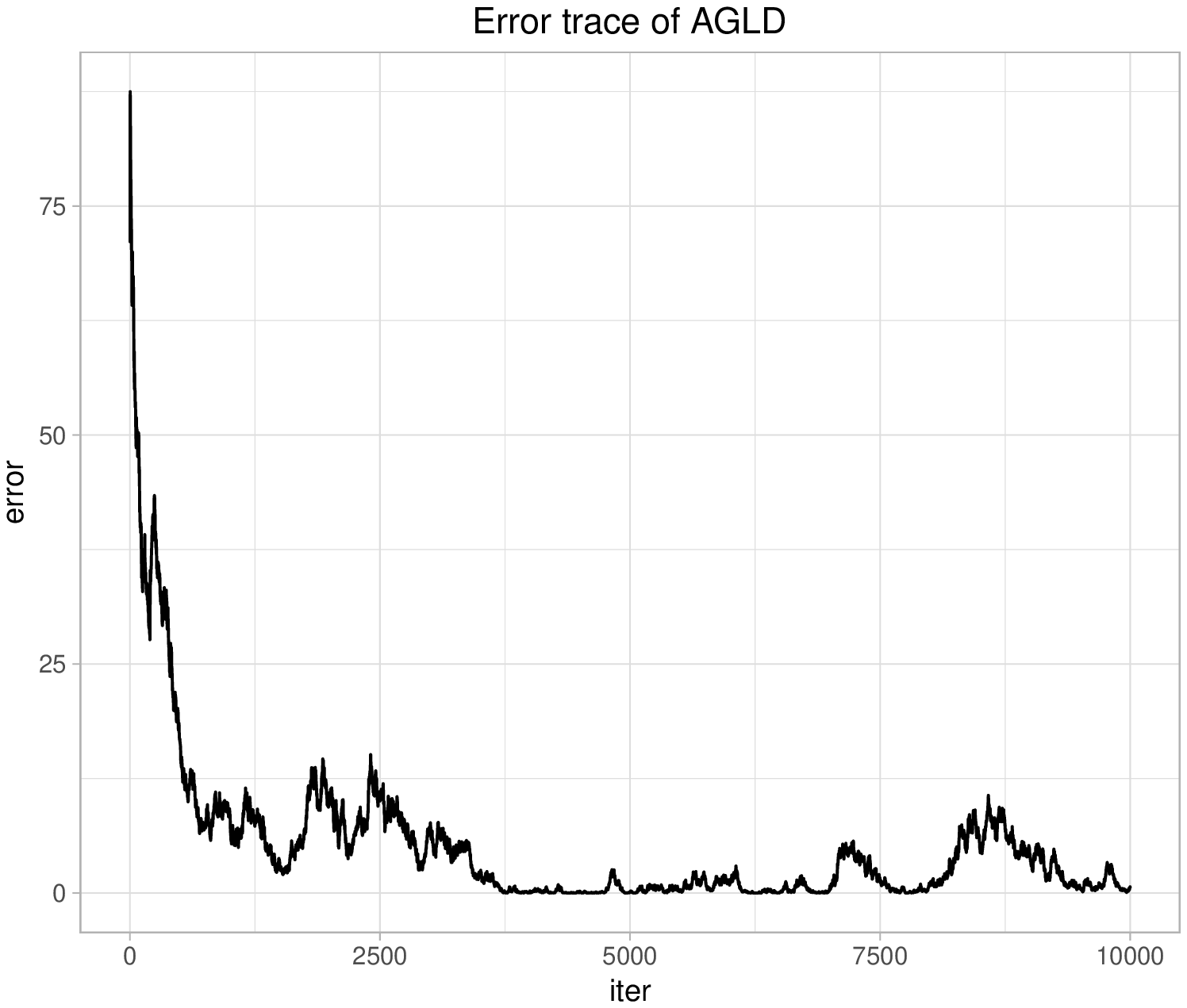}
		\caption{Error for AGLD.}
	\end{subfigure}
	\centering
	\begin{subfigure}[b]{0.4\textwidth}
		\includegraphics[width=\textwidth]{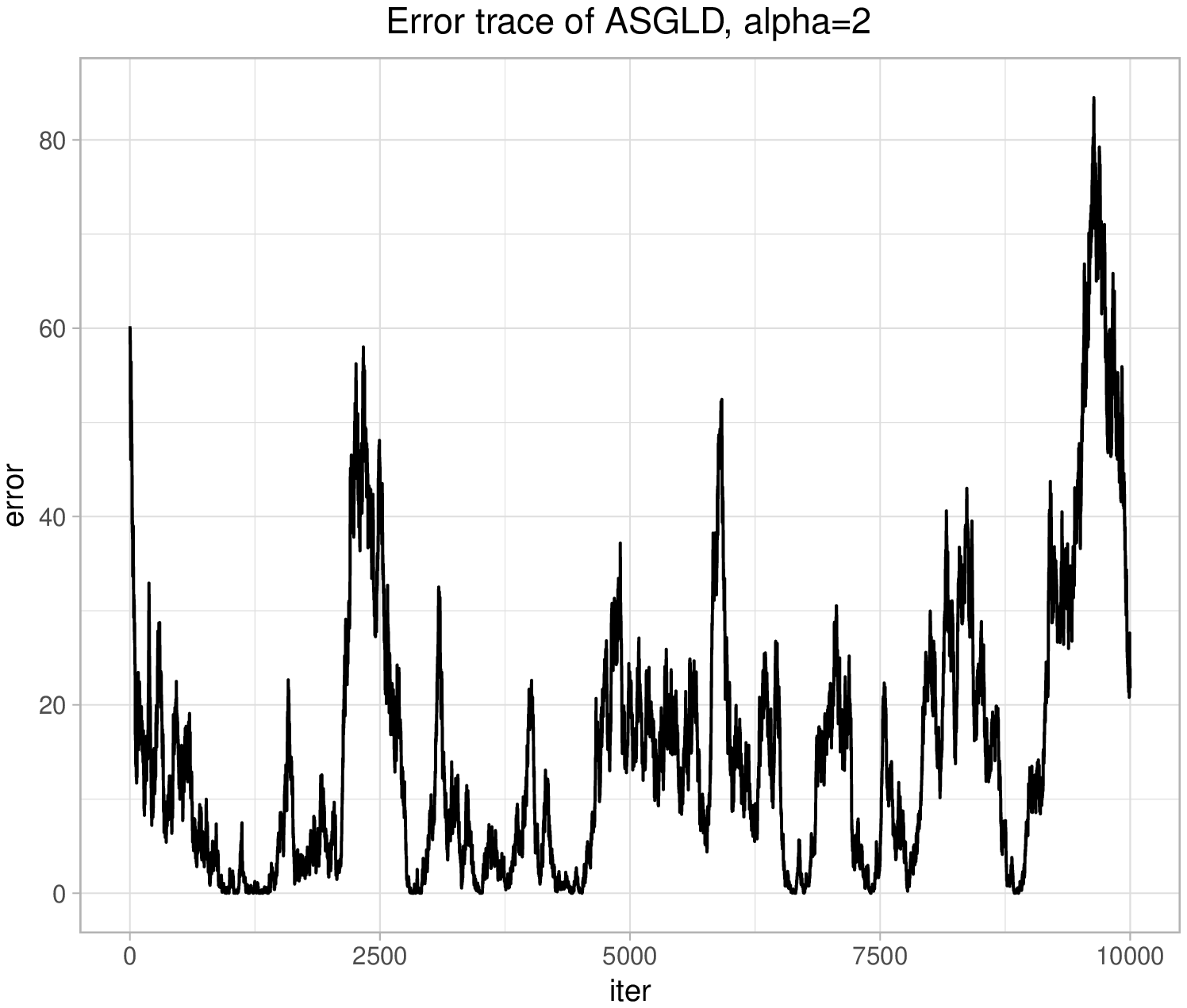}
		\caption{Error for AGLD using $\beta=2$.}
	\end{subfigure}
	~ 
	\begin{subfigure}[b]{0.4\textwidth}
		\includegraphics[width=\textwidth]{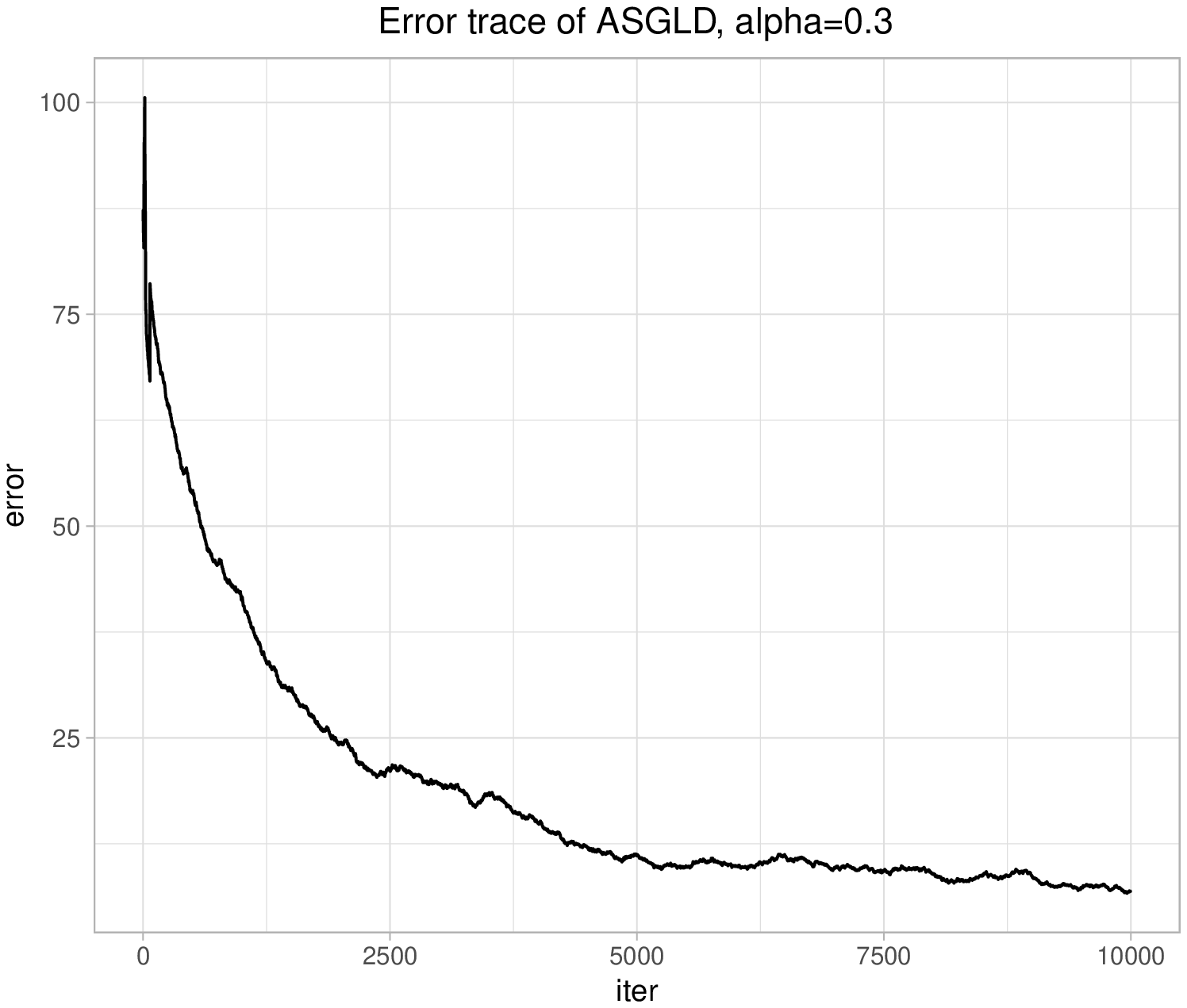}
		\caption{Error for AGLD using $\beta=0.5$.}
	\end{subfigure}
	\caption{Simulation results IV.}\label{fig4}
\end{figure}

The simulation results are presented in Figure \ref{fig11} and,\ref{fig4}.

%
%

Form the simulation results, we can see that SGLD and its variants suffer the convergence issue. Our proposed ASGLD and its variants converges fast for $\beta>\alpha/2$. For $\alpha=1,\beta=0.3$, where we cannot achieve convergence, ASGLD still hits the true value quickly due to large variability from injected noise.  ASG converges stably but slower than ASGLD. AGLD converges even faster than ASGLD, which we have shown in theorem.

\end{document}